\documentclass[letterpaper, 10 pt, conference]{ieeeconf}
\IEEEoverridecommandlockouts
\usepackage{times}
\usepackage[pdftex]{graphicx}
\usepackage{graphics}
\usepackage[cmex10]{amsmath}
\usepackage{url}
\usepackage{threeparttable}
\usepackage{array}
\usepackage{tabularx}
\usepackage{multirow}
\usepackage{multicol}
\usepackage{booktabs}
\usepackage{epstopdf}
\usepackage{rotating}
\usepackage{csquotes}
\usepackage[bookmarks=true]{hyperref}

\usepackage[export]{adjustbox}
\usepackage{amssymb}
\usepackage{amsmath}
\usepackage[all]{xy}    
\usepackage{ifthen}
\usepackage[usenames,dvipsnames]{color}
\usepackage{color, colortbl}
\definecolor{Gray}{gray}{0.9}
\usepackage{enumerate}
\usepackage{bm}
\usepackage{hhline}
\usepackage{xpatch}
\makeatletter
\patchcmd\@makecaption{\\}{.~}{}{\fail}
\makeatletter
\usepackage{algorithm}
\usepackage{algorithmic}
  \usepackage{setspace}
\let\Algorithm\algorithm
\renewcommand\algorithm[1][]{\Algorithm[#1]\setstretch{1.2}}
\usepackage[bottom]{footmisc}

\makeatletter
\let\NAT@parse\undefined
\makeatother
\usepackage[numbers]{natbib}

\newcommand{\secref}[1]{Section~\ref{#1}}

\newcommand{\myparagraph}[1]{\vspace{0.05in}\noindent\textbf{#1}}

\newboolean{draft-mode}
\setboolean{draft-mode}{true}
\newcommand{\sidenote}[1]{\ifthenelse{\boolean{draft-mode}}{\marginpar{\tiny\raggedright\textsf{\hspace{0pt}#1}}}{}}
\DeclareRobustCommand{\arnote}[1]{\ifthenelse{\boolean{draft-mode}}{\textcolor{blue}{\textbf{AR: #1}}}{}}
\DeclareRobustCommand{\mbnote}[1]{\ifthenelse{\boolean{draft-mode}}{\textcolor{cyan}{\textbf{MB: #1}}}{}}

\title{\LARGE \bf Tactile Mapping and Localization from High-Resolution Tactile Imprints}

{\author{Maria Bauza, Oleguer Canal and Alberto Rodriguez\\ 
Mechanical Engineering Department --- Massachusetts Institute of Technology\\
\tt\small <bauza, oleguer, albertor>@mit.edu 
\thanks{This work was supported by the NSF award [IIS-1637753] through the National Robotics Initiative, and by the Toyota Research Institute (TRI). This article solely reflects the opinions and conclusions of its authors and not TRI or any other Toyota entity.
Maria Bauza is the recipient of \emph{La Caixa} Fellowship.}}}

\begin{document}
\maketitle

\begin{abstract}
This work studies the problem of shape reconstruction and object localization using a vision-based tactile sensor, GelSlim.
The main contributions are the recovery of local shapes from contact, an approach to reconstruct the tactile shape of objects from tactile imprints, and an accurate method for object localization of previously reconstructed objects.
The algorithms can be applied to a large variety of 3D objects and provide accurate tactile feedback for in-hand manipulation.

Results show that by exploiting the dense tactile information we can reconstruct the shape of objects with high accuracy and do on-line object identification and localization,  opening the door to reactive manipulation guided by tactile sensing. We  provide  videos and supplemental information in the project's website  \href{http://web.mit.edu/mcube/research/tactile_localization.html}{web.mit.edu/mcube/research/tactile\_localization.html}.

  %
%
\end{abstract}

\section{Introduction}
\label{sec:introduction}

The correlation between hand dexterity and the spatial-and-pressure resolution of its tactile sensors has been of interest for a long time~\cite{jones2006human}.
In the 19th century, Weber explored spatial acuity with the ``two-point touch threshold", i.e., the shortest distance that can be perceived as two separate pressure points. 
Later, Max von Frey studied the sensitivity to different levels of applied pressure~\cite{norrsell1999cutaneous}. 
It comes with no surprise that the regions with finer spatial sensor resolution, and those that are more sensitive to pressure, are the tips of our fingers and the tip of our tongue, both known for their dexterity.

This work builds from a recent interest in image-based tactile sensors such as GelSlim~\cite{donlon2018gelslim} or GelSight~\cite{GelSight_review} which, by virtue of using a soft gel skin and a camera as transducer, achieve very high spatial acuity and pressure sensitivity, yielding highly discriminative tactile signals.
%

\begin{figure}[t]
\centering
	\includegraphics[width=\linewidth]{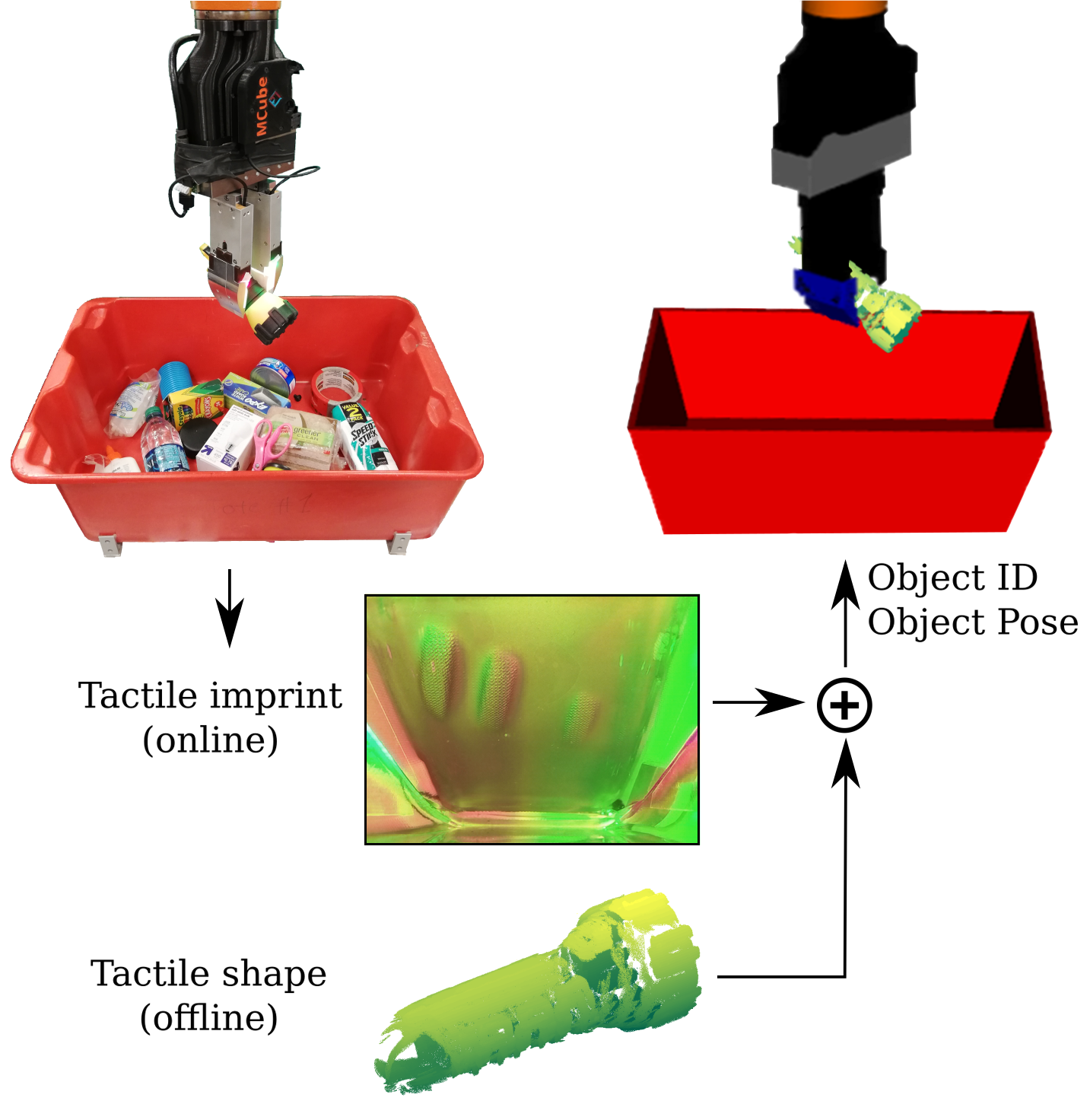}
\centering
\vspace{-7mm}
\caption{\textbf{Tactile mapping and localization.} This work addresses the problem of in-hand object identification and localization using tactile sensing. Given a new tactile imprint from the tactile sensor in the robot's finger, we use the precomputed (offline) tactile map of an object to identify and find its location inside the grasp.} \label{fig:motivation}
\vspace{-5mm}
\end{figure}

In this paper we study the use of tactile imprints as dense descriptors of contact, and demonstrate an approach to reconstructing the tactile shape of an object to facilitate tactile localization (Fig.~\ref{fig:motivation}).
This work is part of an approach to robotic manipulation that has at its core: 
\begin{itemize}
    \item \textbf{Dense tactile descriptors.} High-resolution tactile imprints are dense local descriptors of touch. This opens the door to a large set of classification and regression techniques from machine learning.
    \item \textbf{Crisp tactile memory.} Touching something and recognizing with confidence that it has been touched before. This is essential for data association in estimation/reconstruction problems.
    \item \textbf{Fine differentiation.} Detection of small differences between very similar tactile imprints can be used for part verification or comparison.
    \item \textbf{Contact force distribution.} The deformation of the sensor skin under contact encodes the spatial distribution of internal contact forces. These are key for precise object manipulation~\cite{dong2019,ma2019}. 
\end{itemize}
%
These suggest an approach to manipulation where tactile information does not play a complementary role, but rather a main driver of dexterous manipulation.

In this paper we demonstrate that we can combine tactile imprints with robot kinematics to build a tactile map of an object for localization. To do so, we present 3 contributions: 
\begin{itemize}
\item[1.] Local shape estimation: we use \emph{tactile imprints} to estimate the shape of the contact patch using CNNs. We validate the algorithm in \secref{sec:depth} with known contact geometries that yield sub-millimeter accuracy.
\item[2.] Global tactile mapping: we fuse the tactile imprints and the kinematics (gripper pose and opening) of multiple grasps of a fixed object to reconstruct its global \emph{tactile shape}. This includes the object geometric shape as well as a discrete representation of its tactile imprints. We validate the algorithm in \secref{sec:mapping} by recovering the main dimensions of known objects with an error lower than 5\%.
\item[3.] Object tactile localization: Figure~\ref{fig:motivation} illustrates how we combine tactile imprints with an estimation of the shape of the contact patch to identify and localize a grasped object. Our ICP-based algorithm uses tactile imprints for coarse data association, and contact shape for fine refinement. We validate the process in \secref{sec:localization} with controlled quantitative grasping experiments and in \secref{sec:real} with qualitative results for unstructured picking.
\end{itemize}


\section{Related Work}
\label{sec:relatedWork}

Effective robotic manipulation usually requires a good understanding of the manipulated objects. To this end, many estimation techniques have been proposed to recover object properties or to track and identify them.
With good visibility, visual information can be enough to recover and track most objects \citep{newcombe2011,Salas-Moreno2013,OpenAI2018}. However, in the context of robotic manipulation, occlusions are often unavoidable specially when the robot actively interacts and manipulates objects. To mitigate this problem, some works have explored tactile sensing to better estimate object properties or motion \citep{Bjorkman2013, Allen1999,Ilonen2014,Luo2017, Falco2017}. Most of these works heavily rely on visual information and tend to use tactile information for some form of contact detection.

Recent works such as \citet{Luo2017} and \citet{Falco2017} provide good shape reconstructions or do object localization, but remain restricted to mostly planar objects or use tactile sensors that are bulky and impede dexterous manipulations.
To enable more realistic robotic manipulation, some works have used tactile sensors that are more naturally integrated into robotic arms and hands to explore object properties.

There is extensive work based on tactile sensing that aims at recovering the shape of manipulated objects \citep{Strub2014,Luo2018, Martinez-Hernandez2013,Jamali2016,Yi2016,Driess2017,Sommer2014,zhang2017active,mao2017shape}. Among the most recent works, \citet{Jamali2016} and \citet{Driess2017} actively search for regions that need to be explored based on tactile readings. However, their sensors are low resolution which makes the reconstructed shapes hard to use for other applications. Our approach instead uses a high resolution sensor that not only reconstructs the local contact shape, but reuses their tactile map to enable object identification and localization. We refer the reader to \citet{Luo2017review} for a more extensive review on tactile estimation of objects properties such as shape.

There has also been work on estimating the location of objects using tactile sensing \citep{petrovskaya2011,koval2015,moll2004,pezzementi2011,ottenhaus2016,aggarwal2014}. While some of these works use tactile sensing only as an indicator of contact \citep{koval2015}, others such as \citet{aggarwal2014} or \citet{pezzementi2011} aim to recover both the object motion and its shape. These works are tailored to mostly planar objects and use low resolution tactile sensors. Our approach instead generalizes to 3D objects without relying on prior knowledge.

Our work is also related to the SLAM estimation problem \citep{besl1992,dellaert2006,Yu2015}. The application of SLAM to manipulation of 3D objects however has been challenging because of lack of clean information for data association. Our approach relies on building a dense tactile map of each object offline that enables coarse data association and can later be used for multiple manipulation tasks to both identify objects and locate their pose w.r.t. to the tactile sensor.

In this work we use GelSlim \cite{donlon2018gelslim}, a tactile sensor based on Gelsight~\cite{yuan_2017} that provides high resolution tactile imprints in the form of images. The original Gelsight sensor has been used for object localization of small objects \cite{li_2014}, to complement a vision-based tracker~\cite{izatt_2017}, or recently to recover 3D shapes using also vision and prior shape models~\citep{Wang2018}. However, its design is bulky for practical use in complex manipulation tasks. Instead, GelSlim is integrated in a slim finger that facilitates manipulation~\citep{zeng_2017, Hogan2018}. Leveraging GelSlim's high resolution, we show that our approach can reconstruct tactile maps of objects and use them efficiently to identify and recover their location.

\begin{figure*}[t]
\centering
{\includegraphics[width=\linewidth]{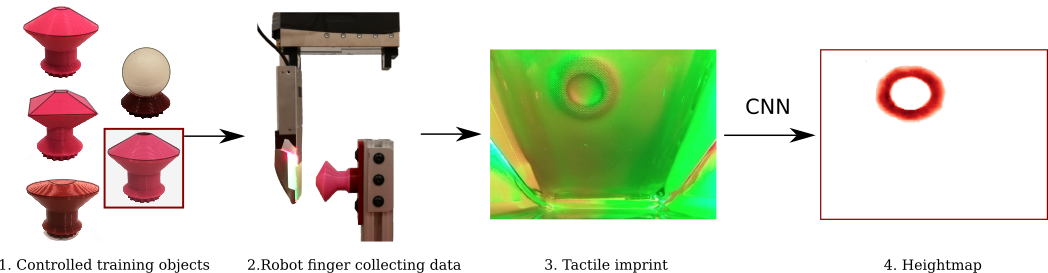}}
\centering
\vspace{-5mm}
\caption{\textbf{Local shape estimation.} To estimate the shape of an object at contact, we built a system that automatically collects data and maps tactile imprints to heightmaps of the local shape. From left to right: a) objects used for training, b) robot collecting data by frontally touching random locations of an object, 3) tactile image recorded during the touch, and 4) heightmap of the object's shape at contact obtained using a trained CNN.} \label{fig:depth_train}
\vspace{-3mm}
\end{figure*}   
\section{Approach: Tactile mapping and localization}
\label{sec:framework}

We present a framework to reconstruct a tactile map of an object to identify and locate it in-hand. Figure~\ref{fig:motivation} illustrates the process where we combine an \emph{off-line} reconstructed tactile shape of a flashlight with an \emph{on-line} imprint to locate its position after being grasped. 
The framework is composed of three steps to address three main challenges:
\begin{itemize}
    \item \textbf{Local shape estimation.} Given a tactile imprint we estimate the local shape of an object during contact. We address this with a self-supervised approach that trains a CNN (Convolutional Neural Net) to map tactile imprints to local contact shapes. The process is described in Section~\ref{sec:depth}.
    \item \textbf{Global tactile and shape mapping.} Given a set of tactile imprints from an object and their respective local shapes, we use the kinematics of the robot to reconstruct a global tactile shape. Section~\ref{sec:mapping} explains this off-line controlled process.
    \item \textbf{Identification and localization.} Given an on-line tactile imprint, we compare it to previously reconstructed tactile shapes to identify the object, and to initialize an ICP-based algorithm to locate it within its global tactile map (Sections~\ref{sec:localization} and~\ref{sec:real}).
\end{itemize}

\section{Local shape estimation} \label{sec:depth}

In this section we explain how to recover the local shape of an object from a tactile imprint. The local shape is given as a heightmap and aims to represent the local geometry of an object at contact. In Fig.~\ref{fig:depth_train}, we use a data-driven approach to build a map between tactile imprints and local shapes.

We automate data collection using a setup that builds upon the system described in \citep{zeng_2017,Hogan2018} which features a robotic arm (ABB IRB 1600id) with a parallel jaw gripper (WSG-50 Weiss) and a GelSlim tactile sensor at each finger \citep{donlon2018gelslim}. In this work, we only use one tactile sensor, but our approach readily extends to both. During data collection, objects are rigidly attached to an external platform (Fig.~\ref{fig:depth_train}) and palpated at different locations and orientations.

To find a map between tactile images and local shapes, we used the 5 controlled geometries in Fig.~\ref{fig:depth_train} to ensure a diverse set of tactile imprints. These objects provide ground truth heightmaps as they produce identifiable local shapes such as the section of a sphere, a semicone, a semipyramid or a semicone that has a cavity at its center (hollow) that can be easily localized in the imprint to facilitate the autonomous labelling of each tactile imprint. 
For each object, we collected 600 pairs of tactile imprints and heightmaps, while holding out 100 for testing.

\myparagraph{Network Architecture.}
Given that tactile images and heightmaps are 2D arrays, we leverage standard CNNs. The basic architecture of the CNN we use is a sequential model of 10 convolutional layers with 64 filters each and a 3-by-3 sized kernel. To improve the robustness to illumination changes, we augment the data including random variations in the 3 channels of the tactile images. We also account for translations by adding two extra channels to the input with the $x$ and $y$ position of each pixel. A more in detail explanation of the data collection and training process can be found in the project's website~\citep{projectwebsite}.

\myparagraph{Evaluation.}
We evaluate the quality of the heightmaps on 100 test images per object. The error reported is the RMSE (Root Mean Squared Error) between the actual and predicted heightmaps w.r.t. to the size of the contact patch.  
To first order, we see that the RMSE decreases by adding more training data and that the average reconstruction accuracy reaches 0.1mm on the test data with only 500 datapoints and 0.060$\pm$0.016mm using 2000 datapoints.

\begin{table}[]
  \small
   \caption{ Cross-validation between objects.}
    \label{tab:cross-validation}
  \centering
    \begin{tabular}{c|c|c|c|c|c|}
     & Sphere & {Cone1} & {Cone 2} & {Hollow} &  {Pyramid} \\ \hline
    All & 0.066 & 0.077 & 0.062 & 0.070 & 0.067  \\ \hline
    Removed & 0.108 & 0.167 & 0.073 & 0.145 &  0.137 \\ \hline
    \end{tabular}
\vspace{-5mm}
\end{table}

\begin{table*}[!t]
  \centering
  \setlength{\tabcolsep}{4 pt}
  \caption{Parameter estimation using shape reconstruction}
  \begin{tabular}{c|c|cc|ccc|cc|cc|}
   {} & Cylinder  & \multicolumn{2}{c|}{Semicone}  & \multicolumn{3}{c|}{Double-cylinder} & \multicolumn{2}{c|}{Cuboid} & \multicolumn{2}{c|}{Semipyramid} \\\hline
  Object  & 
 \hspace{-3mm}
    \begin{minipage}{0.1\textwidth} \centering \vspace{-2mm} \includegraphics[width=\linewidth, keepaspectratio]{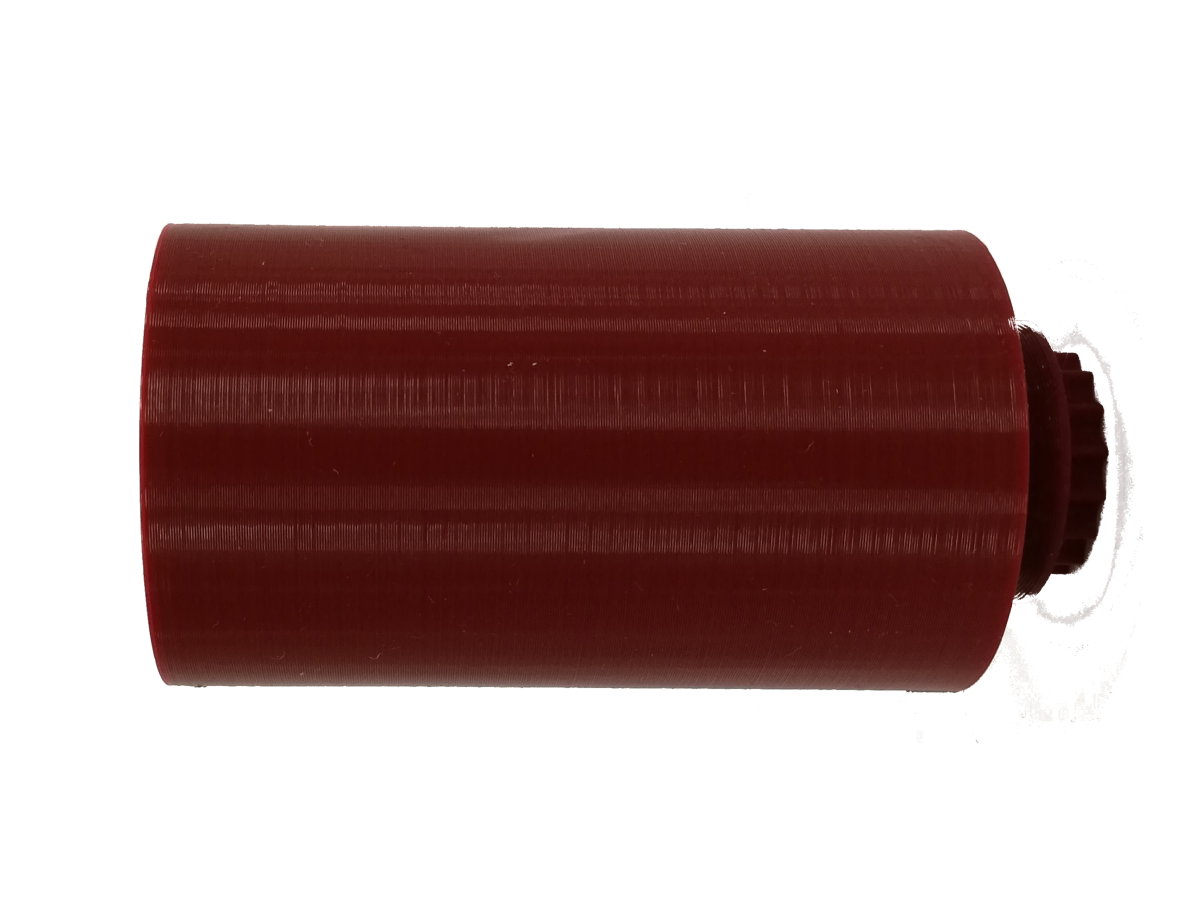} \vspace{-6mm} \end{minipage} \hspace{-3mm}
    &
    \multicolumn{2}{c|}{
    \hspace{-3mm} \begin{minipage}{0.1\textwidth} \centering \vspace{-2mm} \includegraphics[width=\linewidth,  keepaspectratio]{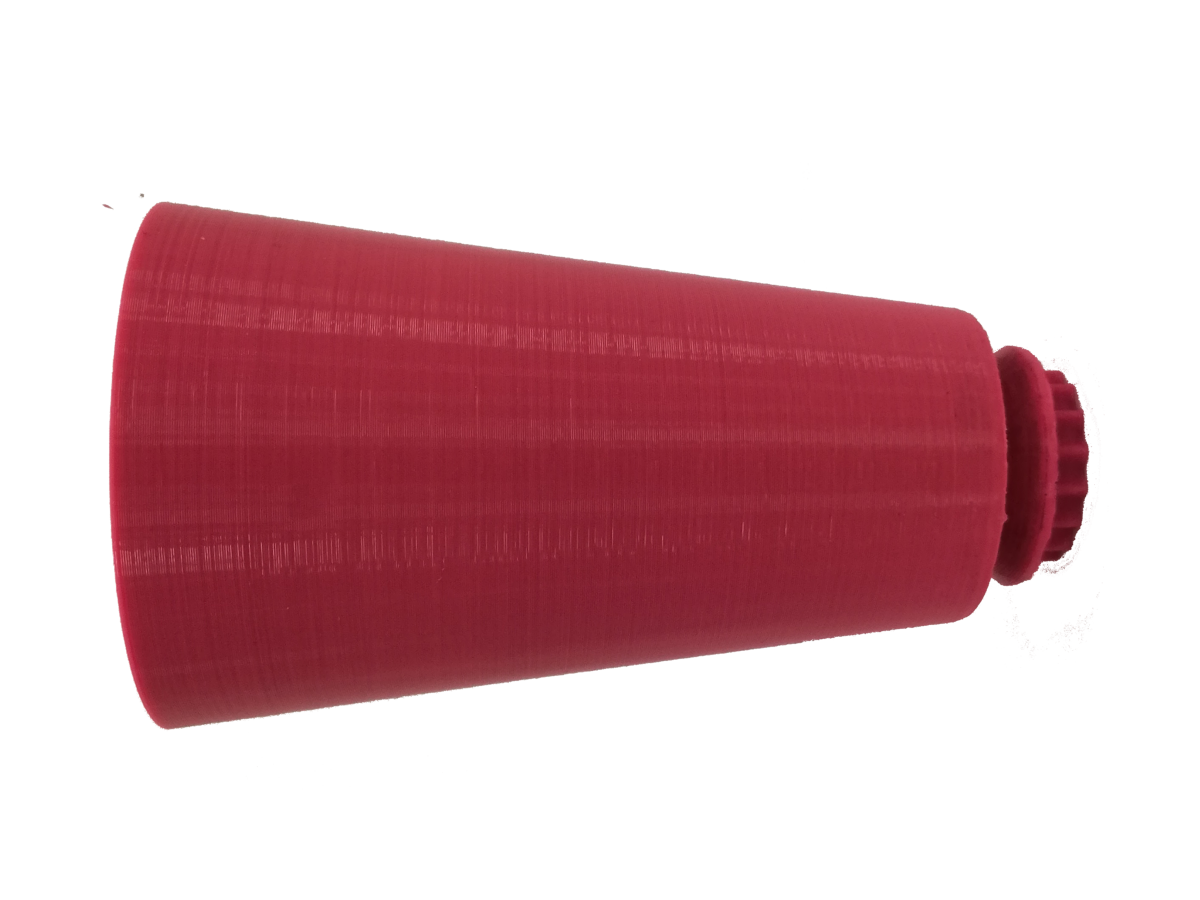} \vspace{-6mm} \end{minipage} \hspace{-3mm}
    }
    &
    \multicolumn{3}{c|}{
    \hspace{-3mm} \begin{minipage}{0.1\textwidth} \centering \vspace{-2mm}
      \includegraphics[width=\linewidth, keepaspectratio]{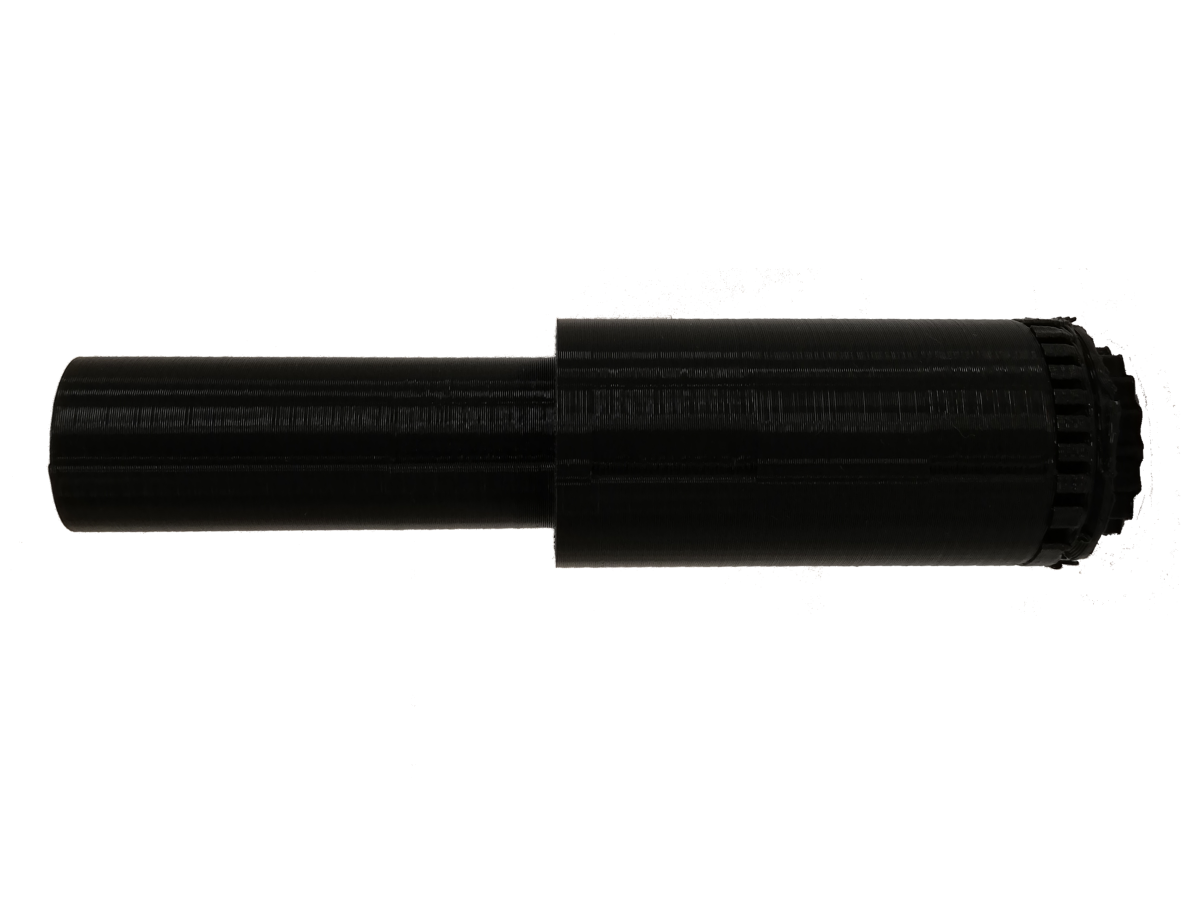}
    \vspace{-6mm} \end{minipage}
        \hspace{-3mm}
    }
    &
    \multicolumn{2}{c|}{
    \hspace{-3mm} \begin{minipage}{0.1\textwidth} \centering \vspace{-2mm} \includegraphics[width=\linewidth,  keepaspectratio]{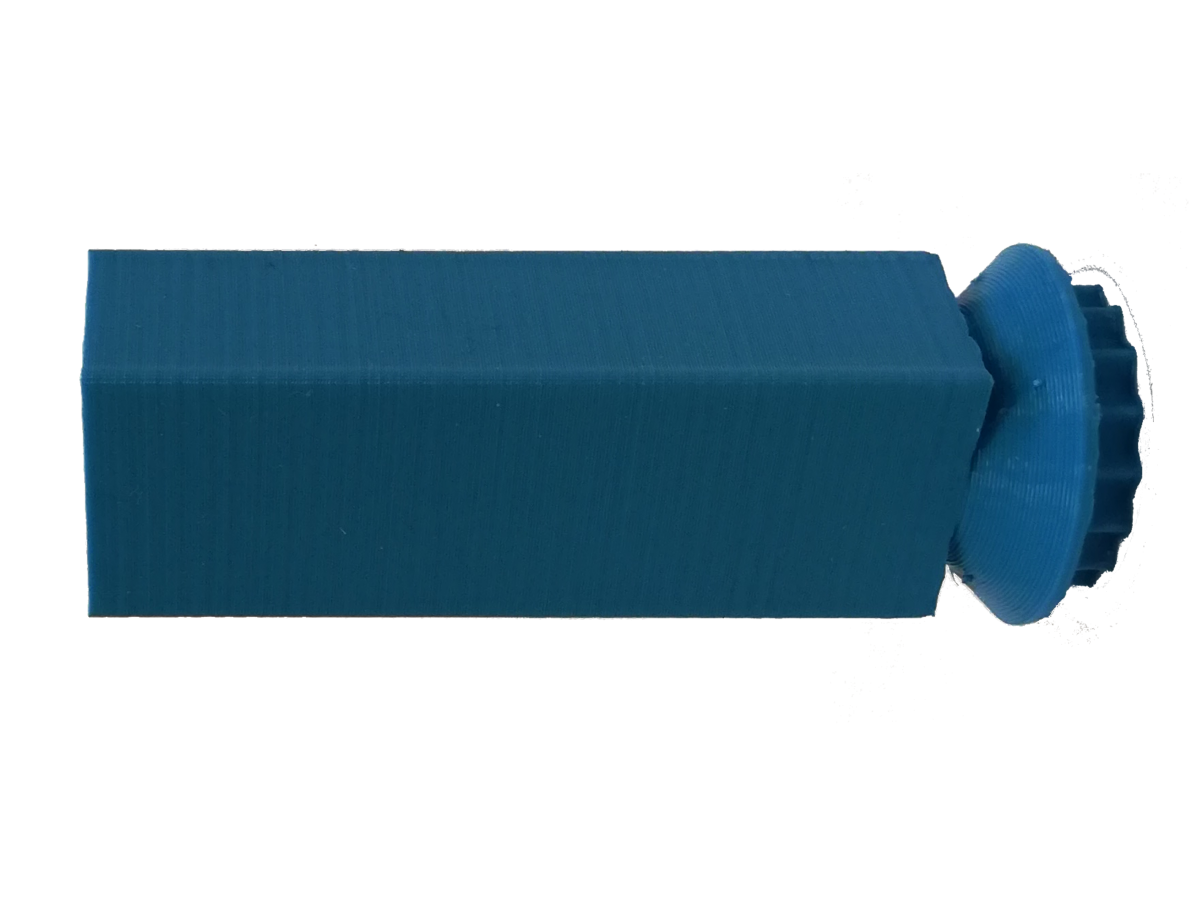} \vspace{-6mm} \end{minipage} \hspace{-3mm}
    }
    &
    \multicolumn{2}{c|}{
    \hspace{-3mm} \begin{minipage}{0.1\textwidth} \centering \vspace{-2mm} \includegraphics[width=\linewidth,  keepaspectratio]{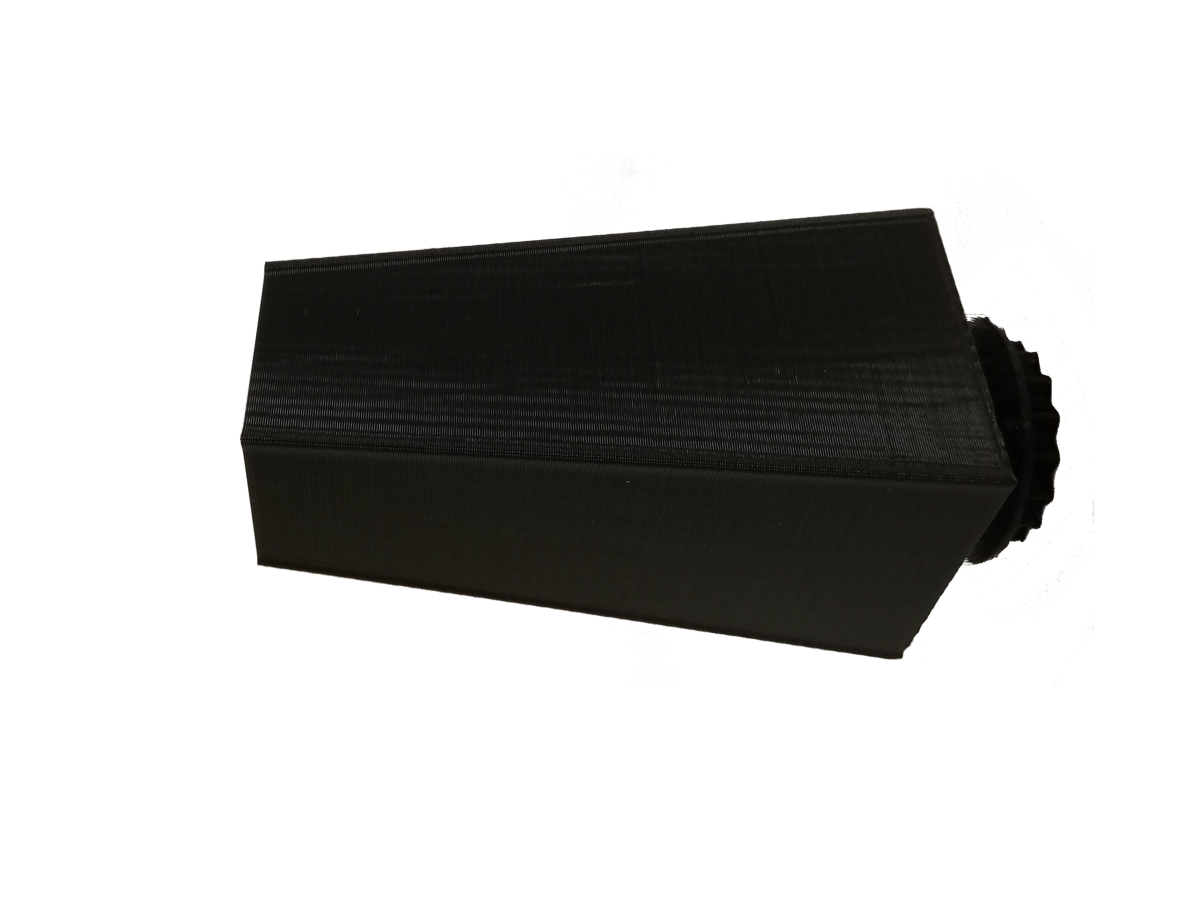} \vspace{-6mm} \end{minipage} \hspace{-3mm}
    }
    \\\hline
    
 Tactile shape  & 
 \hspace{-3mm}
    \begin{minipage}{0.1\textwidth} \centering \vspace{-2mm} \includegraphics[width=\linewidth, keepaspectratio]{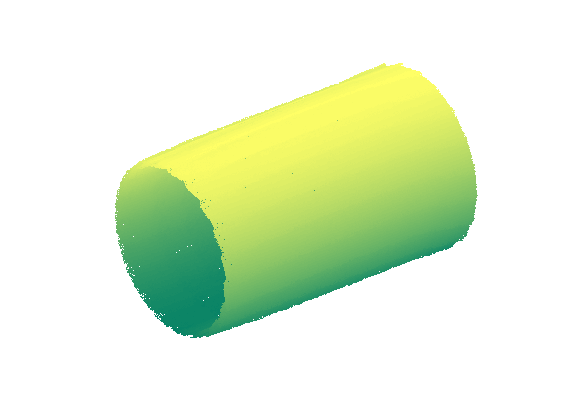} \vspace{-6mm} \end{minipage} \hspace{-3mm}
    &
    \multicolumn{2}{c|}{
    \hspace{-3mm} \begin{minipage}{0.1\textwidth} \centering \vspace{-2mm} \includegraphics[width=\linewidth,  keepaspectratio]{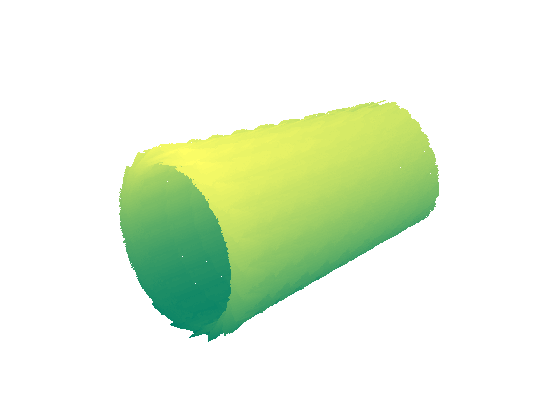} \vspace{-6mm} \end{minipage} \hspace{-3mm}
    }
    &
    \multicolumn{3}{c|}{
    \hspace{-3mm} \begin{minipage}{0.1\textwidth} \centering \vspace{-2mm}
      \includegraphics[width=\linewidth, keepaspectratio]{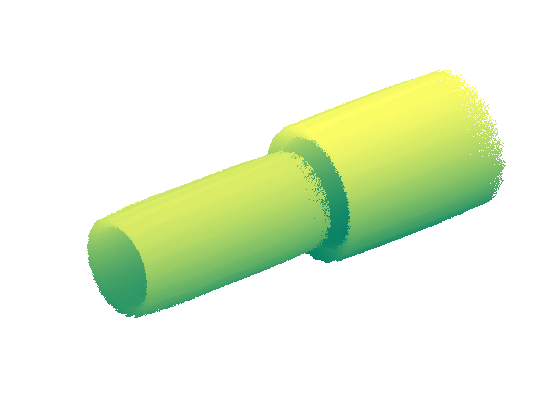}
    \vspace{-6mm} \end{minipage}
        \hspace{-3mm}
    }
    &
    \multicolumn{2}{c|}{
    \hspace{-3mm} \begin{minipage}{0.1\textwidth} \centering \vspace{-2mm} \includegraphics[width=\linewidth,  keepaspectratio]{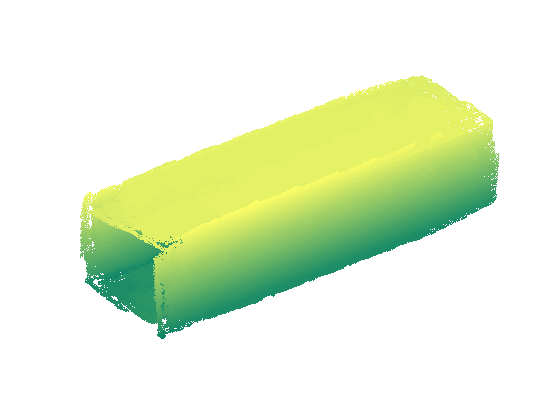} \vspace{-6mm} \end{minipage} \hspace{-3mm}
    }
    &
    \multicolumn{2}{c|}{
    \hspace{-3mm} \begin{minipage}{0.1\textwidth} \centering \vspace{-1mm} \includegraphics[width=\linewidth,  keepaspectratio]{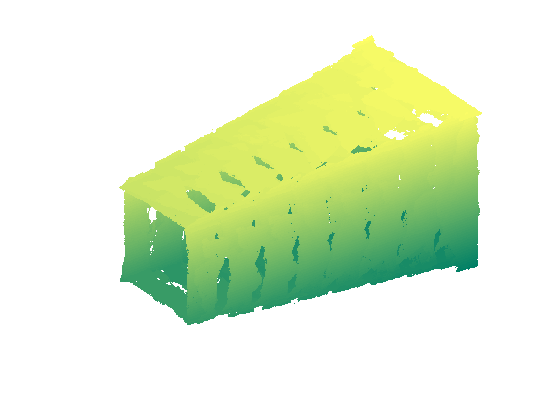} \vspace{-6mm} \end{minipage} \hspace{-3mm}
    }
    \\\hline
    
Parameters & Radius & Base radius &  Slope & Big radius & Small radius & Joint & Side 1 & Side 2 & Base side & Slope \\\hline
    {Real Value}   & 25.0& 20.0 & 7.5$^\circ$ & 16.0 & 11.0 & 60.0 & 25.0 & 20.0  & 20.0 & 7.5$^\circ$ \\\hline
    {Estimation}       &  24.8  & 19.8 & 7.5$^\circ$  & 15.3 & 10.4 & 61.9 & 26.0 & 20.9 & 19.7& 7.2$^\circ$ \\\hline
    {Relative error}       & 2.9\%& 2.2\% & 0.1\% & 4.4\% & 5.5\% & 3.2\%  & 4\% & 4.5\% & 1.5\% & 4\% \\\hline
  \end{tabular}
  \label{table:map_params} \label{fig:mapping} \label{tab:mapping}
 \vspace{-5mm}
\end{table*} 


 Table~\ref{tab:cross-validation} shows the average RMSE (mm) of the heightmap reconstruction on novel shapes by a cross-validation approach. For each training object, we remove it entirely from the training set, train a CNN with 2000 datapoints, and evaluate it both on the whole test set and just on the images from the removed object. Even for hold-out objects, the reconstruction accuracy is on the order of 0.1mm. We observe experimentally that objects like \texttt{semicone1} or \texttt{hollow} that have unique geometric features, such as edges and holes, are more important to the training set. Smaller objects that leave a more reliable imprint on the sensor are also more informative for the training set.

\begin{figure}[t]
\centering
	\includegraphics[width=\linewidth]{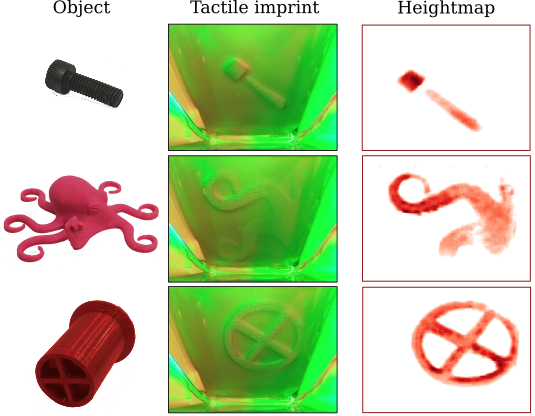}
\centering
\vspace{-5mm}
\caption{\textbf{Examples of local shapes.} Tactile imprints of complex objects and the CNN-computed heightmaps of their local shapes.} \label{fig:depth_test}
\vspace{-5mm}
\end{figure}

Finally, Fig.~\ref{fig:depth_test} shows three examples of how we recover the local shape of novel complex objects. Results indicate that the map built to translate tactile imprints to heightmaps is accurate at recovering the local geometry of objects and can be extended to reconstruct their global shapes.

\section{Global Tactile And Shape Mapping}\label{sec:mapping}

This section describes and analyzes how to combine a set of tactile imprints from an object to recover its tactile shape. This method for shape recovery relies on the accuracy of the robot kinematics and the gripper, and the precision of the heightmaps described in the previous section.

We first explain how each tactile imprint of the tactile map can be localized in the world frame. From the heightmap, we construct a point cloud in the sensor's frame using an accurate calibration of the intrinsic parameters of the sensor's camera. We get this calibration by probing with the tactile sensor on known points w.r.t.  the robot base. Then we localize the point cloud in the world's frame by assuming a rigid and calibrated transformation between sensor, gripper and robot arm. Finally, we stitch all the point clouds obtained from the same object by adding them together into a single point cloud. The project's website~\cite{projectwebsite} provides a more detailed description of these steps.

To test this approach, we reconstructed the tactile shapes of the controlled objects in table~\ref{tab:mapping}. The data collection process is similar to the one in Section~\ref{sec:depth} where we fix the position and orientation of each object, and grasp it along many locations and orientations while recording the tactile imprints. For uniformity of resolution, the grasps follow an equispaced grid of 10mm of resolution along the planes defined by the gripper orientation w.r.t. the fixed object. We considered 3 orientations for the grasp: 0 and $\pm$20$^\circ$ in the vertical axis to promote diversity in the tactile imprints.  

We evaluate the accuracy of the tactile maps by estimating the main dimensions of each object. We see in table~\ref{tab:mapping} that the error in most parameters is less than a few millimeters. From our experience, a denser grid of tactile imprints improves the accuracy of the tactile map for some objects at the cost of longer exploration times. We opted for keeping the exploration equal for all the analyzed tactile maps to enable fair comparison among objects.  

\section{Tactile Localization}\label{sec:localization}  

\begin{figure*}[t]
\centering
{\includegraphics[width=\linewidth]{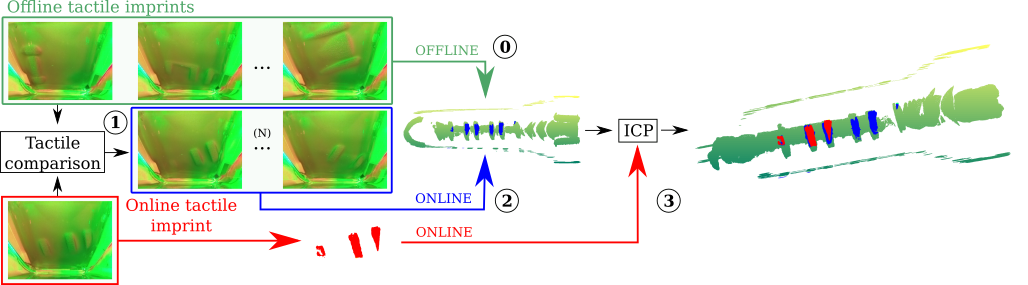}}
\centering
\vspace{-6mm}
\caption{\textbf{Tactile localization using CTI-ICP-N.} Given a new tactile imprint, we find its location in an offline computed tactile map (0) by following the steps: (1) find the N touches from the map that are more similar to the new one, (2) create an auxiliary point cloud with these N touches that is a subset of the global one, and (3) use ICP to stitch the local point cloud from the new tactile imprint to the auxiliary one to locate its pose in the global shape. } \label{fig:localization}
\vspace{-2mm}
\end{figure*} 

Given the tactile map of an object, our goal is to effectively use it for robotic manipulation. To show this, we examine and evaluate how to localize the object based on correspondences between the tactile shape and local tactile imprints. 

The proposed approach is described  in Fig.~\ref{fig:localization}. We first recover the local shape in the sensor's frame as a point cloud using its tactile imprint as in Section~\ref{sec:depth}. Then we stitch this local point cloud to the global tactile shape of the object and infer how the resulting point cloud is located w.r.t to the tactile sensor. Finally, given the robot kinematics and the gripper opening we estimate the actual pose of the object in the world frame. We consider 3 algorithms to stitch the local point cloud: RANDOM, CTI, and our proposed approach CTI-ICP-N which uses CTI to provide a coarse approximation of the object pose and ICP to refine it.

\myparagraph{RANDOM.} Assumes that the sensor's pose during contact is the same as one of the poses (randomly selected) associated with the tactile imprints used to build the global shape. This approach is naive but sets a baseline to assess the performance of other methods.

\myparagraph{CTI.} The pose of the tactile sensor for the new imprint corresponds with the \textit{closest tactile imprint} (CTI) used to construct the global shape. To compute the similarity between touches, we first map each tactile image to a feature vector using as encoder the predictions of a ResNet50 trained on ImageNet~\citep{he2016} without its last fully connected layers. Next we measure the cosine distance between images and only consider those whose gripper opening differs in less than 2mm from the current gripper opening. As a result, the new touch is stitched to the global shape as if it was recorded at the same location as its closest tactile imprint.

\myparagraph{CTI-ICP-N} This approach combines CTI with ICP (Iterative Closest Point) as described in Fig.~\ref{fig:localization}. We start by finding the N closest images to the new tactile imprint, using the same metric as CTI. Then we build an auxiliary point cloud that includes the shapes from all those N closest imprints of the global shape. Finally we do ICP to relocate the local point cloud (initially at the pose of the closest imprint) w.r.t to the auxiliary point cloud to better estimate its pose. 

The case where N is the total number of tactile imprints is equivalent to doing ICP between the local point cloud and the global one, but in practice does not provide better results than considering small values of N and it is more time consuming. We hypothesize this is because of the sparsity and size of the global shape that can be confusing to ICP. By cropping the global shape we indirectly provide focus to the ICP's exploration which reduces the chance that it will get stuck in a local minimum. Figure~\ref{fig:localization} exemplifies how using low values of N prioritizes comparisons of the new point cloud with regions that are more similar to it making the approach more desirable both in terms of accuracy and computation.

\begin{table*}[!t]
  \caption{Localization error for different objects}
  \centering
  \setlength{\tabcolsep}{4 pt}
  \begin{threeparttable}
  \begin{tabular}{c|c|c|c|c|c|}
        & Scissors  & Tape  & Brush & Flashlight & Gum Box \\\hline
        
Object  & 
    \hspace{-3mm}
    \begin{minipage}{0.1\textwidth}
    \centering \includegraphics[width=1.1\linewidth, keepaspectratio]{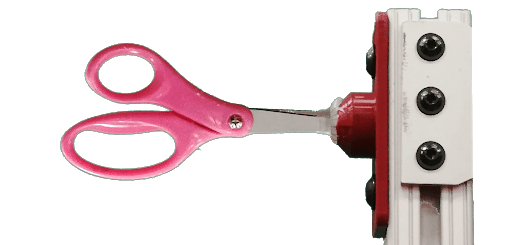}
     \end{minipage}
    \hspace{-3mm} 
    &
    \hspace{-3mm}
    \begin{minipage}{0.1\textwidth}
    \centering 
        \includegraphics[width=1.1\linewidth,  keepaspectratio]{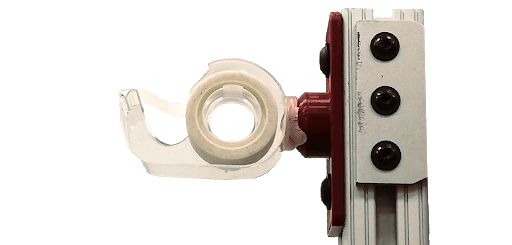}
    \end{minipage}
        \hspace{-3mm}
    &
    \hspace{-3mm}
    \begin{minipage}{0.1\textwidth}
    \centering 
      \includegraphics[width=1.2\linewidth, keepaspectratio]{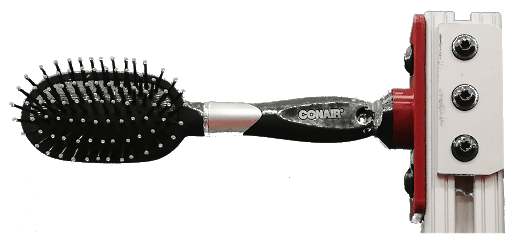}
    \end{minipage}
        \hspace{0mm}
    &
    \hspace{-3mm}
    \begin{minipage}{0.1\textwidth}
    \centering
        \includegraphics[width=1.1\linewidth, keepaspectratio]{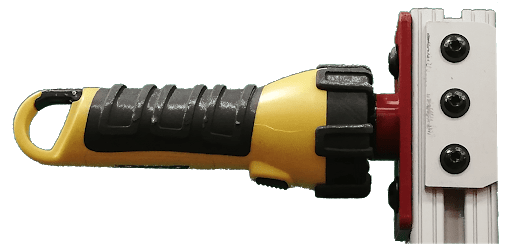}
    \end{minipage}
        \hspace{-3mm} 
    &
    \hspace{-3mm}
    \begin{minipage}{0.1\textwidth}
    \centering 
        \includegraphics[width=1.1\linewidth,keepaspectratio]{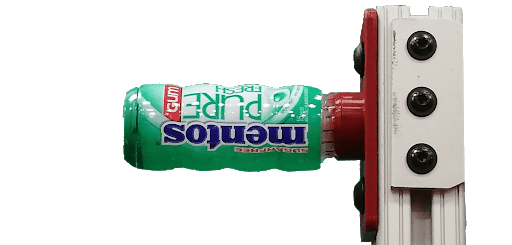}
    \end{minipage}
        \hspace{-3mm} 
    \\\hline
    
Tactile shape & 
 \hspace{0mm}
    \begin{minipage}{0.1\textwidth}
    \centering
         \includegraphics[width=0.95\linewidth, keepaspectratio]{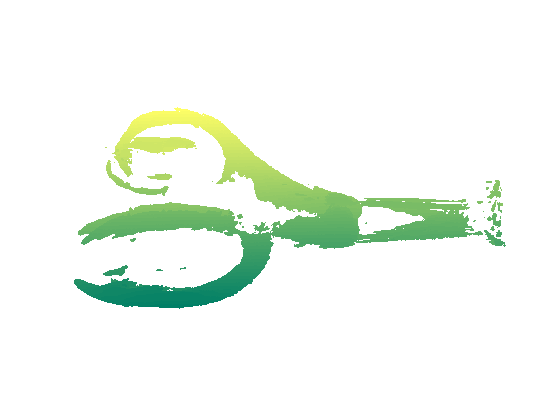}
    \end{minipage}
    \hspace{0mm} 
    &
    \hspace{0mm}
    \begin{minipage}{0.1\textwidth}
    \centering
        \includegraphics[width=0.95\linewidth, keepaspectratio]{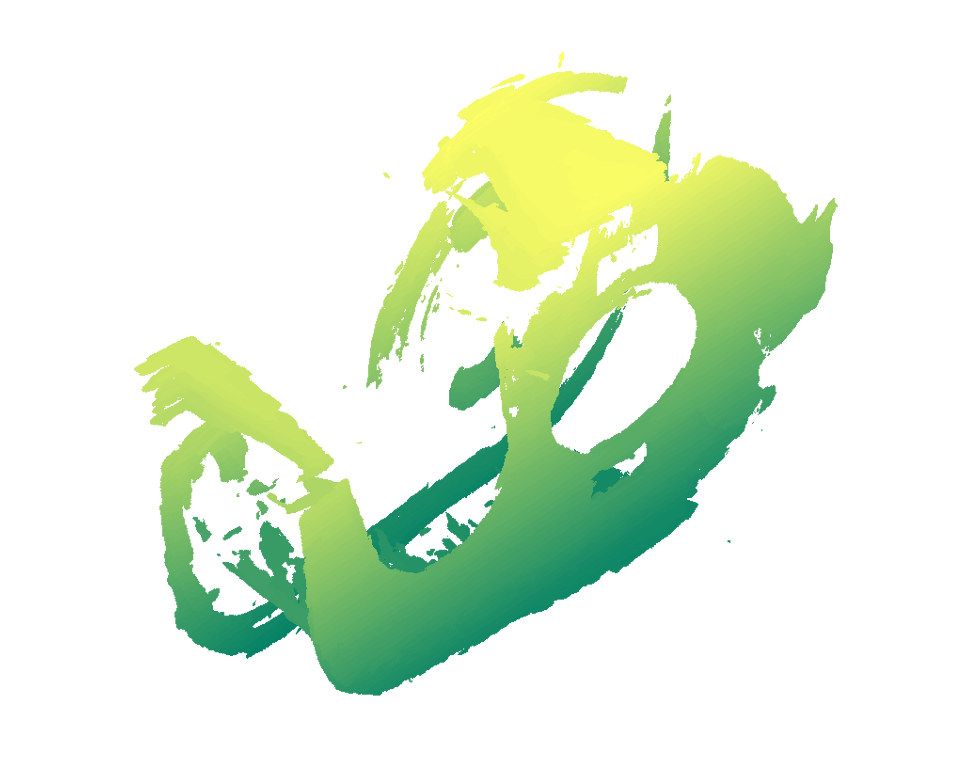}
    \end{minipage}
        \hspace{0mm}
    &
    \hspace{0mm}
    \begin{minipage}{0.1\textwidth}
    \centering
      \includegraphics[width=0.95\linewidth, keepaspectratio]{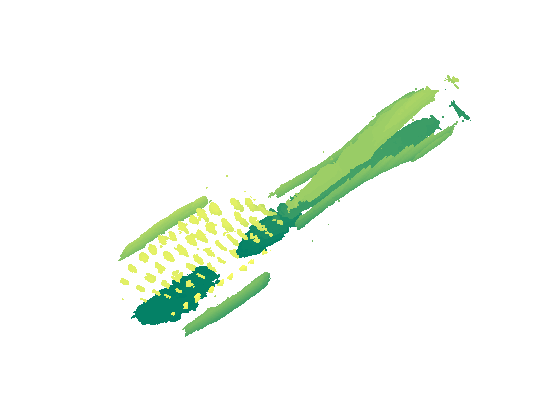}
    \end{minipage}
        \hspace{0mm}
    &
    \hspace{0mm}
    \begin{minipage}{0.1\textwidth}
    \centering
        \includegraphics[width=\linewidth, keepaspectratio]{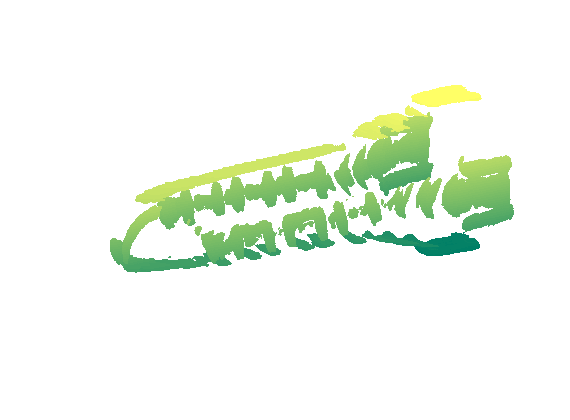}
    \end{minipage}
        \hspace{0mm}
    &
    \hspace{0mm}
    \begin{minipage}{0.1\textwidth}
    \centering
        \includegraphics[width=\linewidth, keepaspectratio]{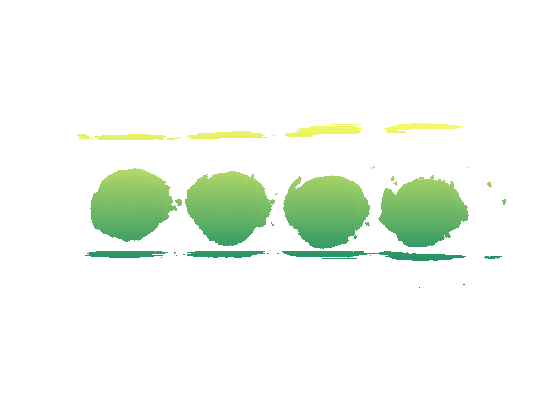}
     \end{minipage}
        \hspace{0mm}
    
    \\\hline
    
    \textbf{RANDOM} & 46.4 & 38.4 & 51.3 & 62.2 & 36.1\\\hline
    \textbf{CTI}   &  10.0& 10.0& 10.0 & 20.0  & 20.0 \\\hline
    \textbf{CTI-ICP-1}       & 7.7 & 7.5  & 9.7 & 22.1 & 20.5 \\\hline
    \textbf{CTI-ICP-5}       & 6.4 & 6.1  & 10.0 & 22.6 & 20.4 \\\hline
    
    \textbf{RMSE distribution} & 
 \hspace{0mm}
    \begin{minipage}{0.1\textwidth}
    \centering
        \includegraphics[width=\linewidth,trim={0 -2mm 0 -2mm}]{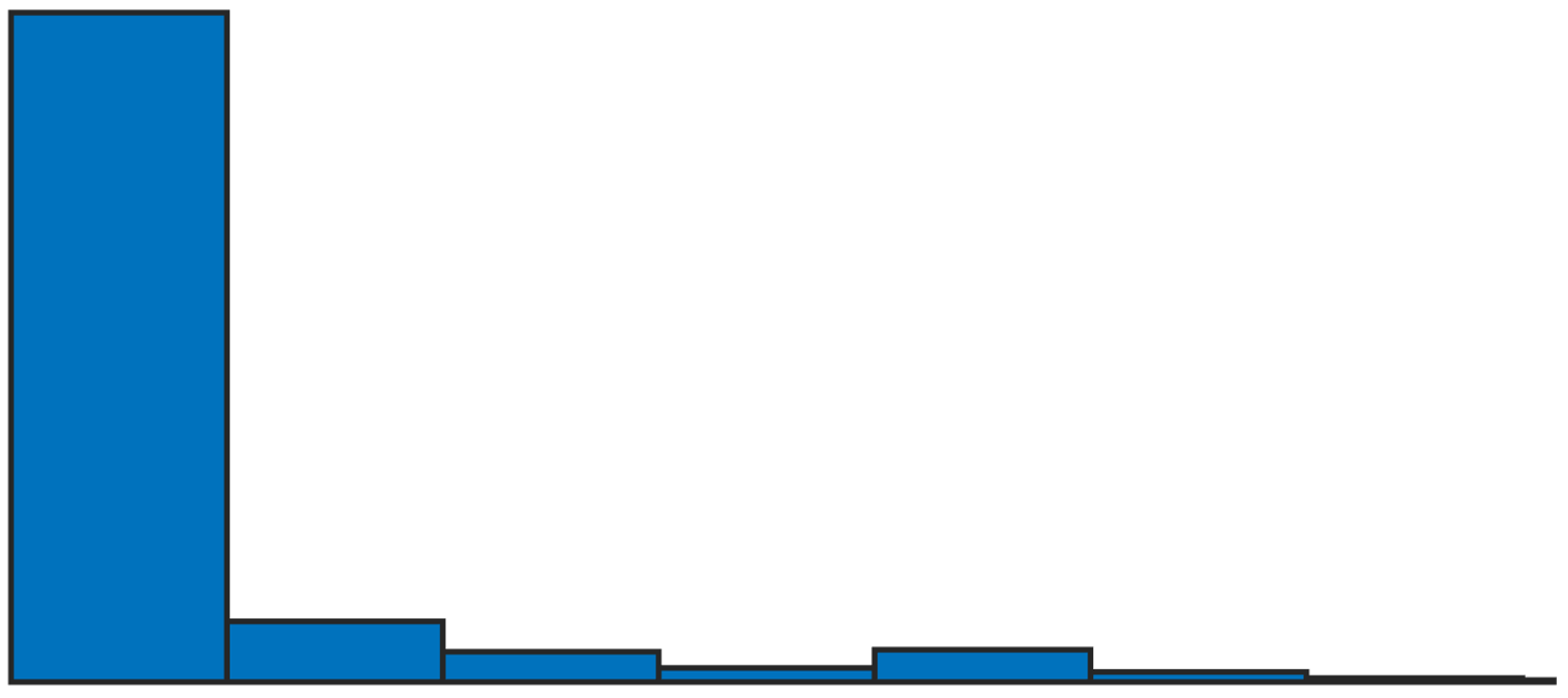}
    \end{minipage}
    \hspace{0mm} 
    &
    \hspace{0mm}
    \begin{minipage}{0.1\textwidth}
    \centering
        \includegraphics[width=\linewidth]{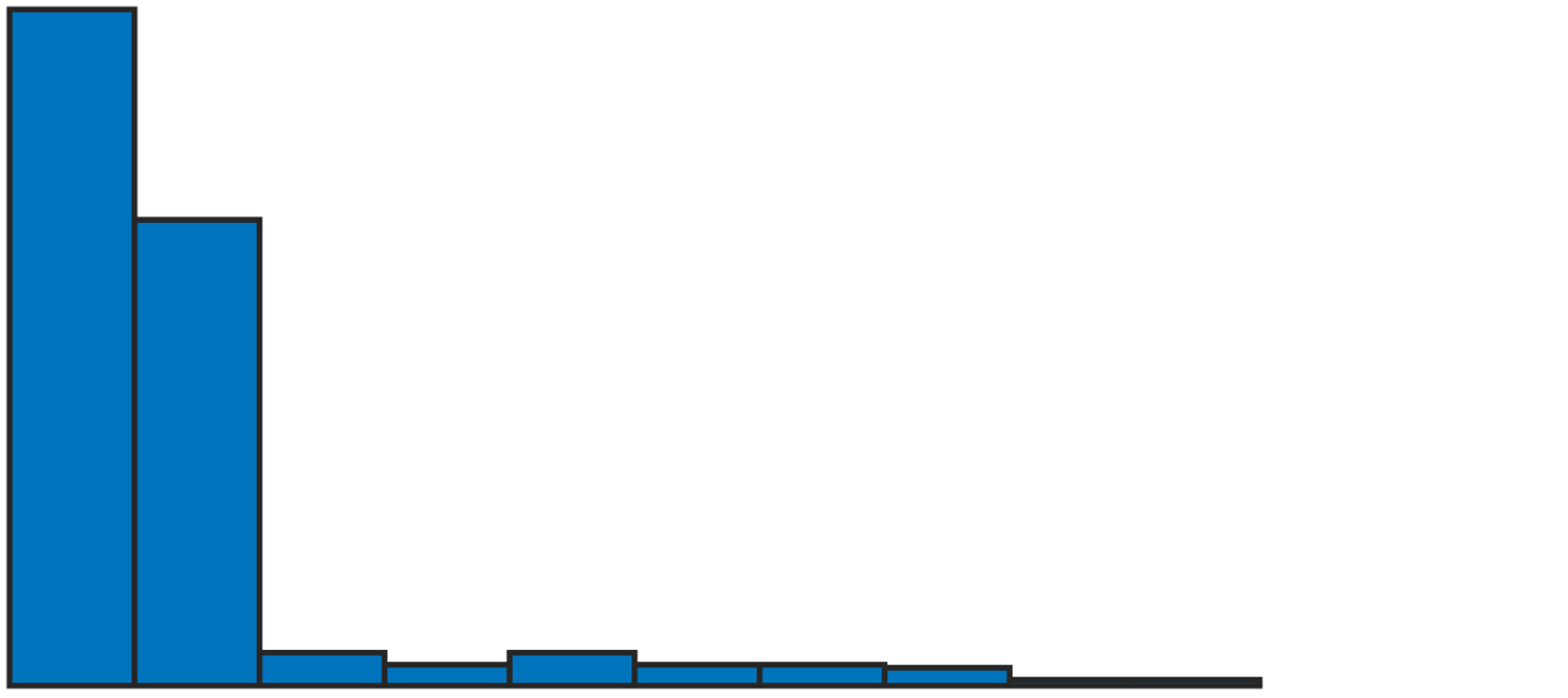}
    \end{minipage}
        \hspace{0mm}
    &
    \hspace{0mm}
    \begin{minipage}{0.1\textwidth}
    \centering
      \includegraphics[width=\linewidth, keepaspectratio]{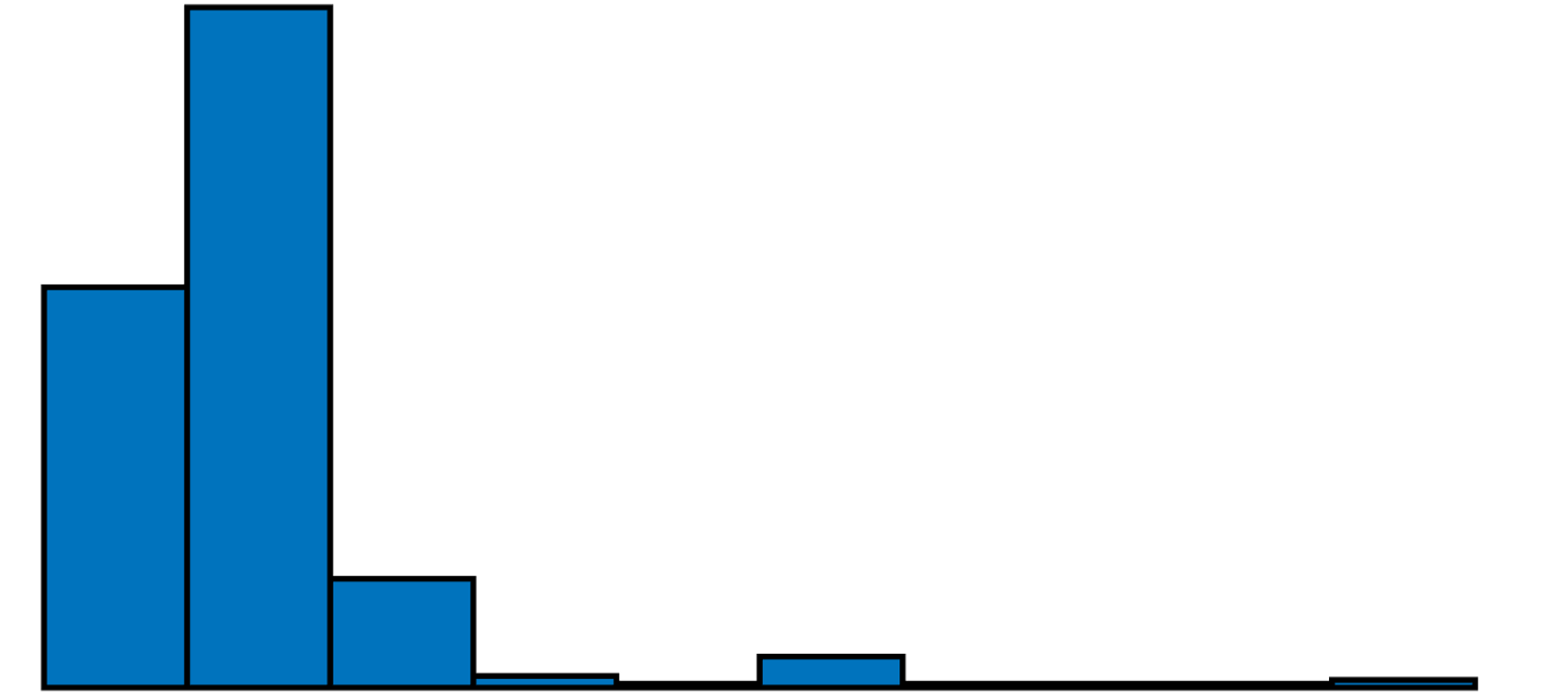}
    \end{minipage}
        \hspace{0mm}
    &
    \hspace{0mm}
    \begin{minipage}{0.1\textwidth}
    \centering
        \includegraphics[width=\linewidth]{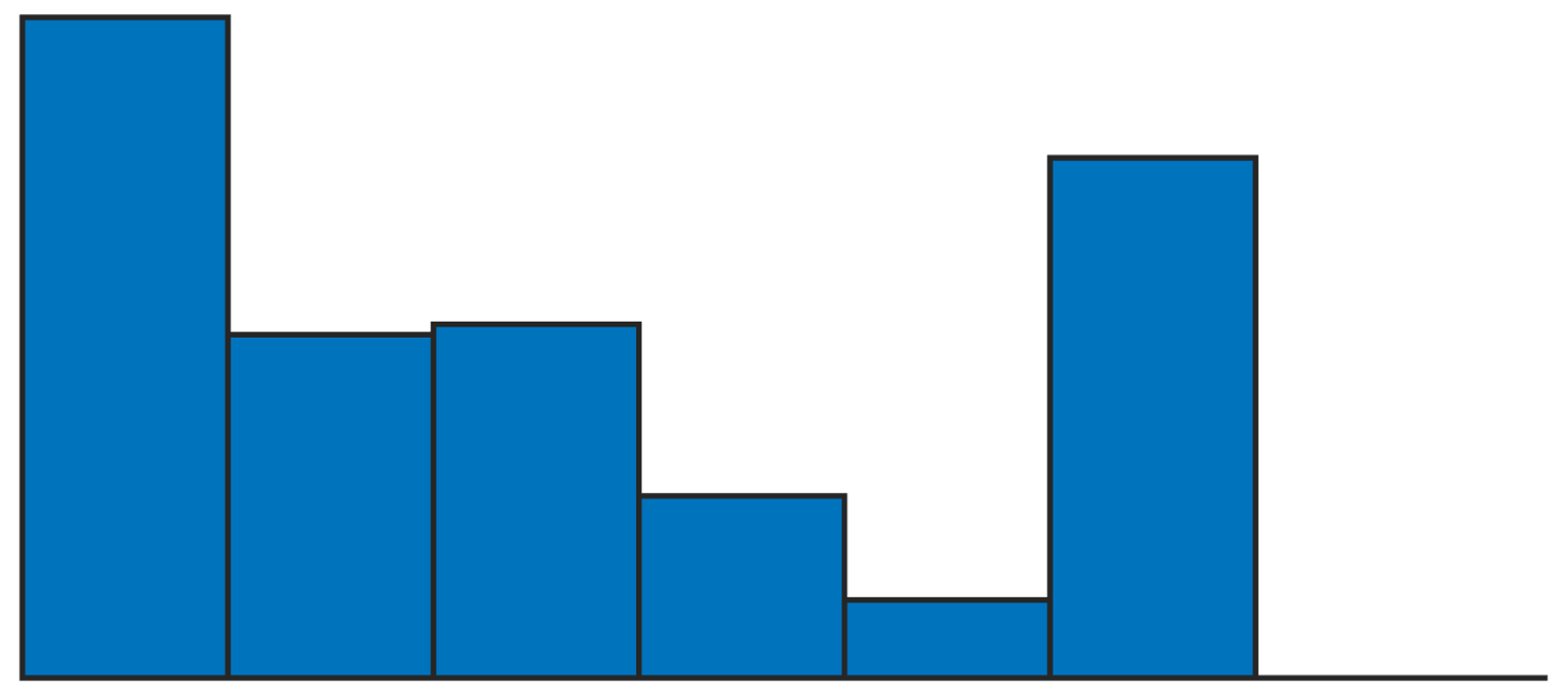}
    \end{minipage}
        \hspace{0mm}
    &
    \hspace{0mm}
    \begin{minipage}{0.1\textwidth}
    \centering
        \includegraphics[width=\linewidth]{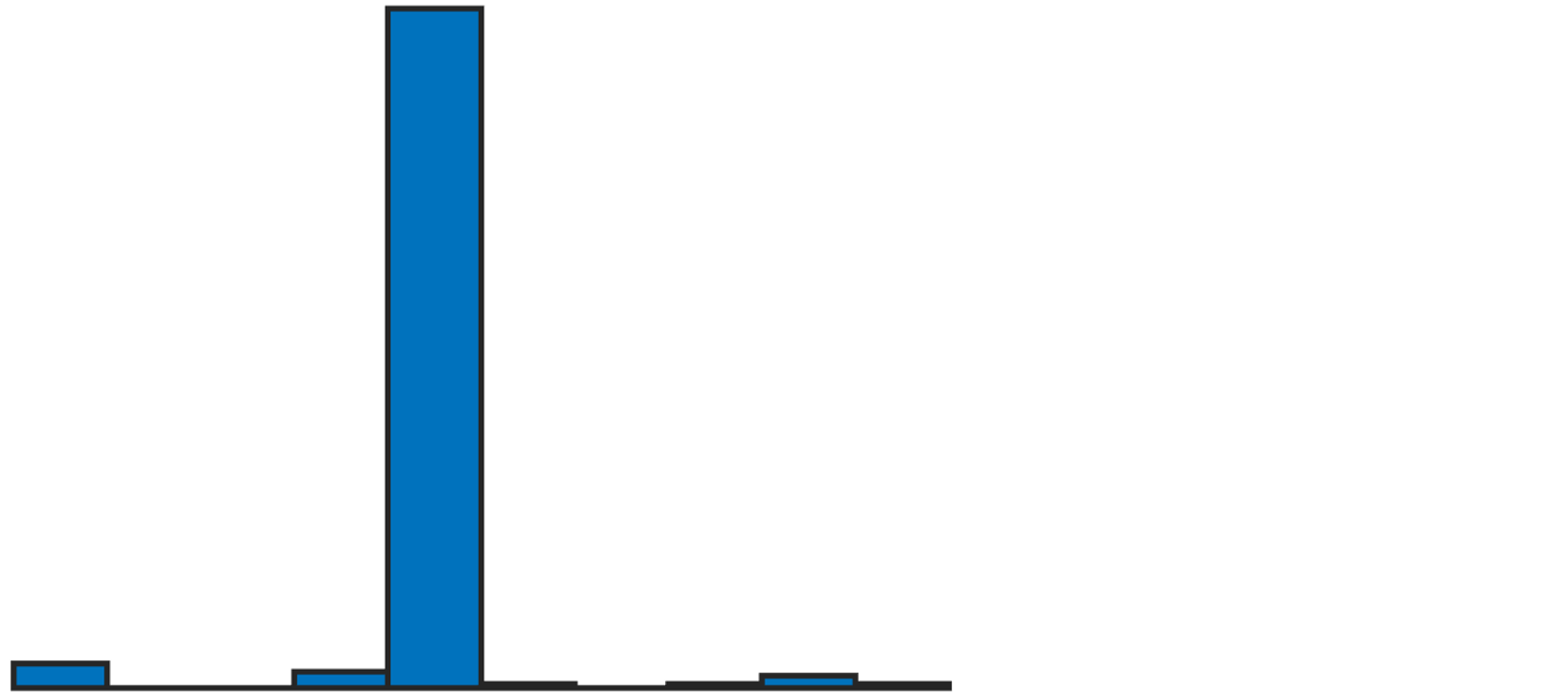}
     \end{minipage}
        \hspace{0mm}
    \\\hline
      
  \end{tabular}
\begin{tablenotes}\item[] \footnotesize {Median of the RMSEs (mm) and RMSE histograms between 0 and 80mm for CTI-ICP-5.}
\end{tablenotes}
\vspace{-5mm}
\end{threeparttable}
  
  \label{table:regrasp_accuracy}
\end{table*} 

\myparagraph{Evaluation.} To evaluate the approach, we considered five common objects and built their global maps as in Section~\ref{sec:mapping} by rotating them 0, 90, 180 and 270$^\circ$. Table~\ref{table:regrasp_accuracy} shows the objects and their reconstructed tactile shapes. The objects vary in several dimensions including weight, length, width, symmetries and texture. For instance, the gum box is small and symmetric while the brush is longer and exhibits different tactile imprints depending on the grasp. 

We use a cross-validation approach to evaluate the accuracy of the algorithms. For each object, we remove one of the touches used to build its global shape and aim to localize it back. We measure the error in the relocation by computing the RMSE between the original point cloud and its final position after estimating its pose. Results in table~\ref{table:regrasp_accuracy} show the median values of the RMSEs after removing each local point cloud once from the global tactile map. CTI, i.e., finding the closest tactile imprint to the removed one, has high accuracy as the median RMSE is on the order of 10mm for 3 of the 5 objects and 20mm for the other two. This is encouraging because the distance between grasps for the same object orientation is at least 10mm. For the gum box, we observed that even with a more dense grid it is difficult to achieve an error lower than 20mm because of its symmetries: tactile imprints separated by 20mm look identical at plain sight which makes their local shapes very similar. 

\begin{figure}[t]
\centering
	\includegraphics[width=\linewidth]{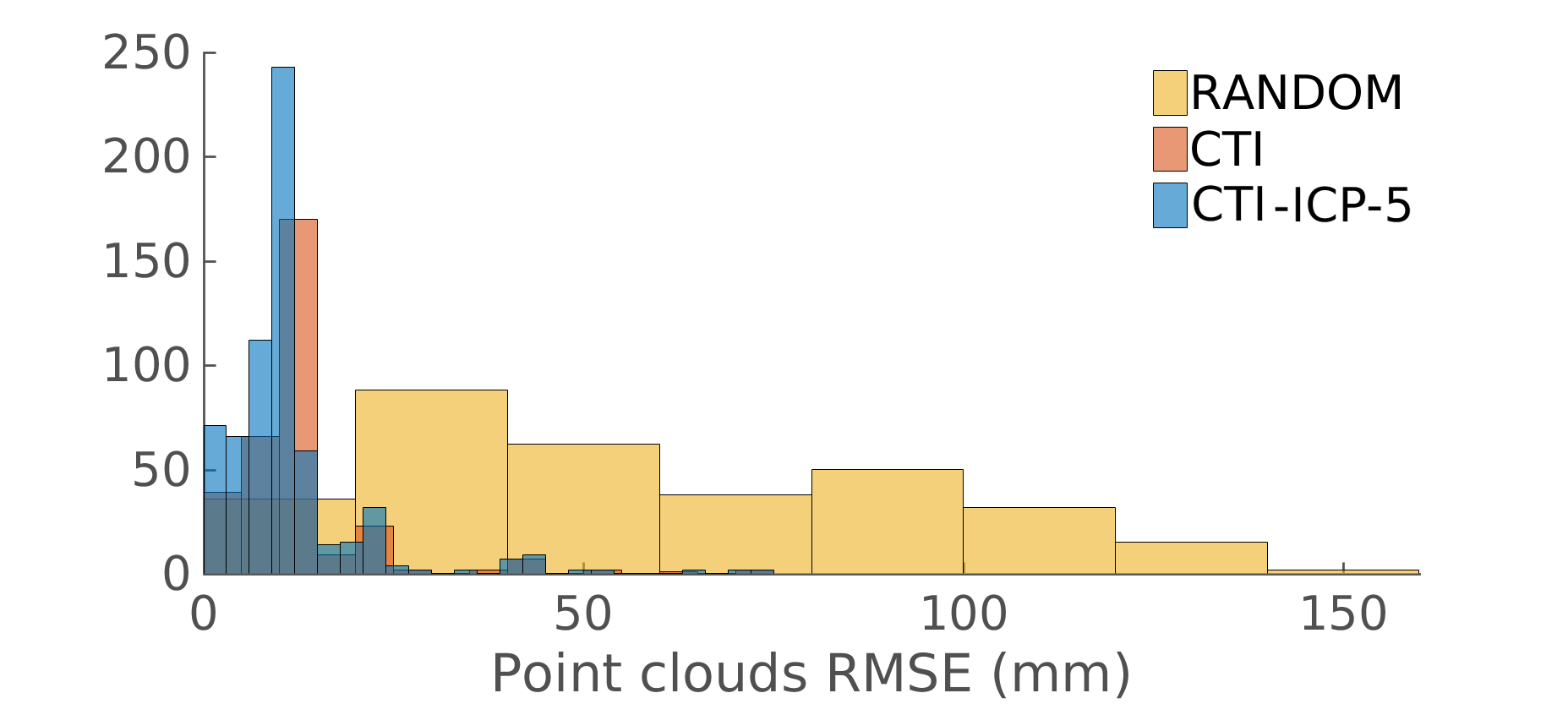}
\centering
\vspace{-6mm}
\caption{\textbf{Distribution of errors.} We compare the distribution of the RMSEs from the hair brush for RANDOM, CTI and CTI-ICP-5 (CTI-ICP-1 was in between CTI and CTI-ICP-5). While the distribution is spread for RANDOM, both CTI and CTI-ICP-5 produce a great improvement on the prediction of the point cloud pose. CTI and CTI-ICP-5 still suffer from outliers making their medians almost the same, but CTI-ICP-5 has more density to the left than CTI. This suggests that CTI-ICP-5 can refine its predictions w.r.t. CTI and be more accurate when the initial guess is good. }
\label{fig:localization_histogram}
\vspace{-8mm}
\end{figure}

\begin{figure}[t]
\centering
	\includegraphics[width=\linewidth]{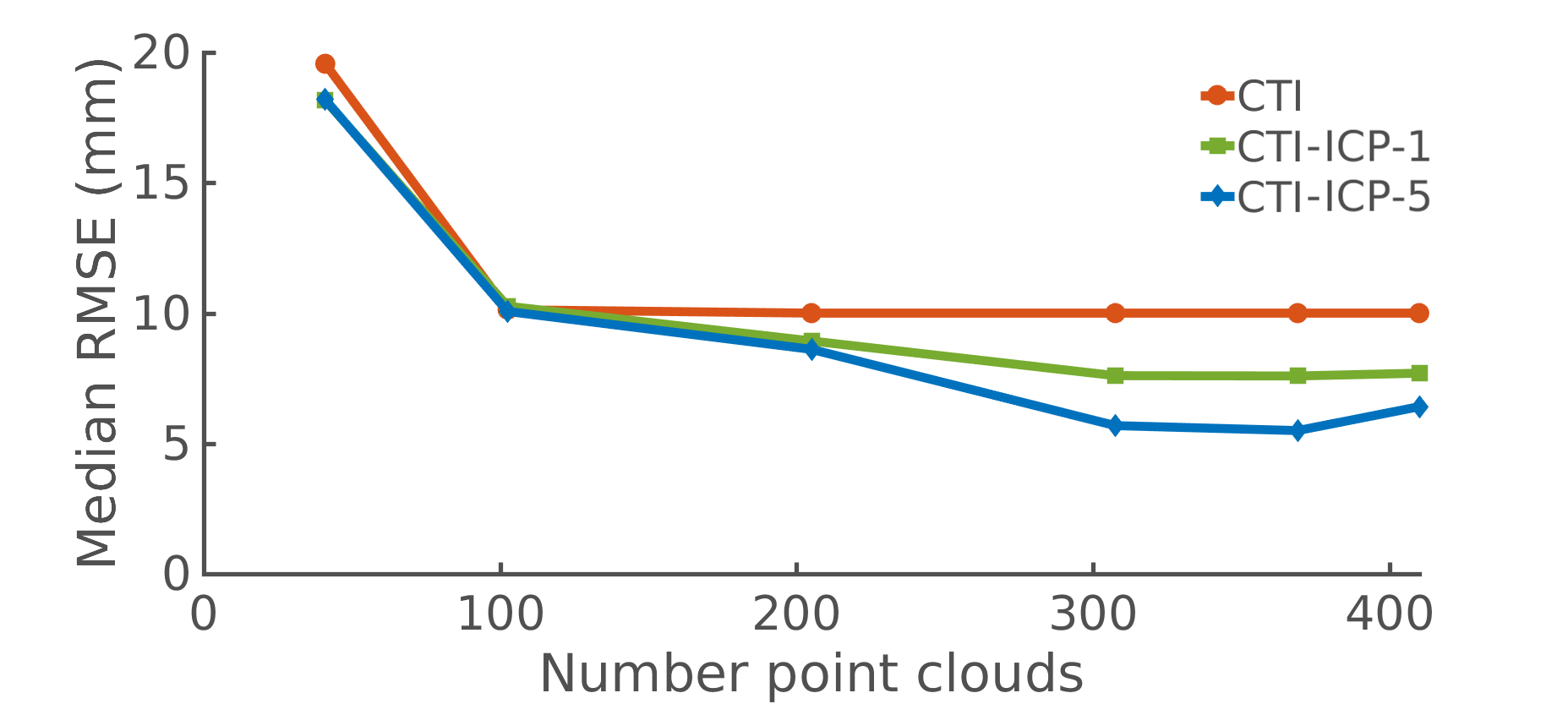}
\centering
\vspace{-6mm}
\caption{\textbf{Error vs. number of point clouds} We study the median of RMSEs for the scissors depending on the number of tactile imprints used to build the tactile map. With only 10\% (around 50) of the point clouds, the median is around 20mm, but decreases to 10mm if we consider a 25\% (around 100) of point clouds or more. As the number of point clouds increases, CTI does not improve, but CTI-ICP-1 and CTI-ICP-5 reduce the error by almost half by making local adjustments to the pose of the point cloud.} \label{fig:localization_results}
\vspace{-2mm}
\end{figure}  

From the results, we also conclude that for objects with a wide distribution of contact patches, doing ICP can boost the localization accuracy. The median of RMSEs for the scissors and the tape already shows that applying CTI-ICP-5 leads to lower RMSEs than CTI. Figure~\ref{fig:localization_histogram} compares the distribution of RMSEs for RANDOM, CTI and CTI-ICP-5 for the hair brush. As expected, the distribution of errors is tighter and closer to zero for CTI and CTI-ICP-5. However, we observe that CTI-ICP-5 leads to a greater improvement in those cases where there is an initial good guess as its distribution is more titled to the left compared to CTI. While the improvements of CTI-ICP-5 against CTI can not be seen directly through the median of the RMSEs, the histograms show that CTI-ICP-5 accurately refines the pose of some local point clouds.

\myparagraph{Size of the tactile shape.} We study how the number of tactile imprints used to build the tactile shape affects the accuracy. To tackle this question, we built several global maps for the scissors using only a percentage (10, 25, 50 or 75\%) of the touches and evaluated the accuracy when stitching one of the discarded touches to the incomplete tactile maps. Figure~\ref{fig:localization_results} shows the RMSE median for CTI, CTI-ICP-1 and CTI-ICP-5 depending on the number of point clouds. As expected, adding more point clouds helps reduce the error in localization. However, after considering a 25\% percent of the point clouds the median RMSE for CTI stays at 10mm. In comparison, CTI-ICP-N approaches improve their accuracy with more touches as they adjust the position of the stitched point cloud reducing almost by half the error.

\section{In-hand identification and localization} \label{sec:real}


Given a set of tactile shapes from explored objects and a new tactile imprint, our goal is to recover both the identity of the object and its location in-hand. We identify it by comparing the new tactile imprint to the ones used to create the global maps of each object, and assigning the identity from the most similar tactile imprint. Once identified, we stitch the new tactile imprint to its global shape as in Section~\ref{sec:localization} and obtain its pose w.r.t. to the tactile sensor.

\begin{figure}[t]
\centering
	\includegraphics[width=\linewidth]{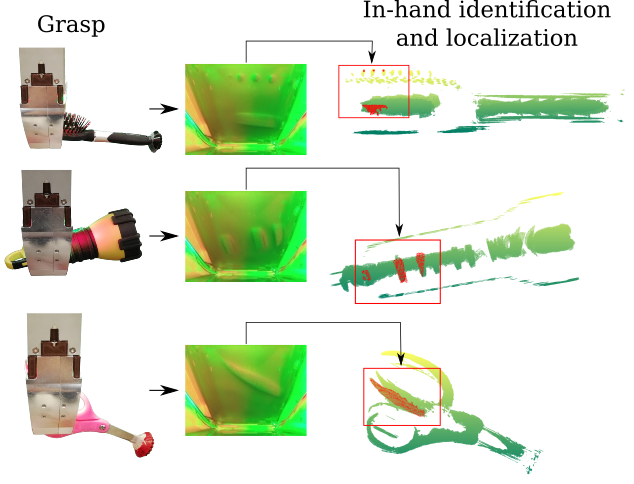}
\centering
\vspace{-8mm}
\caption{\textbf{Object identification and localization.} From each random grasp and tactile imprint, our approach identifies correctly the object and accurately estimates its position in-hand. The objects are identified among those in table~\ref{table:regrasp_accuracy} and localized using the tactile maps from that table.} \label{fig:grasps}
\vspace{-5mm}
\end{figure}

Figure~\ref{fig:grasps} shows 3 examples of grasped objects and their tactile imprints. Our approach correctly identifies each object and estimates its location in-hand. The solution is fast enough to provide real time estimations of the pose as the only steps are: one CNN pass, a similarity comparison in a feature vector space and ICP with small point clouds.

\section{Discussion and Future work} 
\label{sec:discussion}

This paper presents an approach to build a tactile shape and use it to identify and localize an object in-hand. Given a tactile imprint, we learn an accurate model of the local shape at contact. By combining several tactile imprints we build a tactile map of an object and use it for localization. We compare several algorithms for object location given a new tactile imprint and show that the accuracy of the estimates is tightly related to the resolution of the tactile shape, but can be improved combining CTI for coarse localization and ICP for fine refinement. As a result, this approach yields accurate tactile feedback that can be used in real manipulation tasks. 

While our approach has the potential to significantly enhance in-hand manipulation through tactile feedback, there are many opportunities for improvement:

\myparagraph{Using visual information.} The approach presented is purely tactile-based, but that answers to the desire to prove a point: tactile is an enabler of manipulation. A practical system would also exploit visual feedback to create more complete maps. We believe this is possible as our tactile maps provide an estimation of the global shape in the form of point cloud. 

\myparagraph{Considering prior information}. For many tasks, object information is already available offline. Using shape priors or object properties such as mass or texture can help our approach produce a more accurate tactile map. 

\myparagraph{Dealing with hard-to-localize objects.} Some objects are harder or impossible to accurately localize with a single tactile imprint because of their symmetries or lack of texture. In these situations, our approach can still provide a set of possible poses for the object. The combination of multiple imprints could help to better estimate its actual pose.

\myparagraph{Using multiple tactile sensors.} We only use one tactile sensor, but extending to multiples is relatively straightforward. For a two-jaw gripper, two sensors would likely make our approach more discriminative and accurate at location.

\myparagraph{Handling negative information.} Our high resolution sensor does not only provide information about where contact happens, but also where it does not. However, our current solution does not use this information despite we believe adding these constraints could greatly enhance performance.

\myparagraph{Improving tactile similarity.} Super-sizing tactile data collection can provide many more tactile imprints and allow to train a better feature space to discriminate between them. 

\myparagraph{Improving stitching technique.} Using ICP on the point clouds from the tactile imprints has proven more challenging than what we expected. We believe there are ways to mitigate this issue by choosing the right metrics to deal with the sparsity of tactile point clouds.

To make sure our approach is robust and useful for complex manipulation tasks, our next steps include addressing many of the points discussed above. We believe our method provides a reliable and accurate source of tactile feedback that will open a lot of research possibilities within a tactile approach to dexterous manipulation.  

\bibliographystyle{IEEEtranN} 
{\footnotesize \bibliography{mb-icra19-shape}} 

\begin{thebibliography}{41}
\providecommand{\natexlab}[1]{#1}
\providecommand{\url}[1]{#1}
\csname url@samestyle\endcsname
\providecommand{\newblock}{\relax}
\providecommand{\bibinfo}[2]{#2}
\providecommand{\BIBentrySTDinterwordspacing}{\spaceskip=0pt\relax}
\providecommand{\BIBentryALTinterwordstretchfactor}{4}
\providecommand{\BIBentryALTinterwordspacing}{\spaceskip=\fontdimen2\font plus
\BIBentryALTinterwordstretchfactor\fontdimen3\font minus
  \fontdimen4\font\relax}
\providecommand{\BIBforeignlanguage}[2]{{%
\expandafter\ifx\csname l@#1\endcsname\relax
\typeout{** WARNING: IEEEtranN.bst: No hyphenation pattern has been}%
\typeout{** loaded for the language `#1'. Using the pattern for}%
\typeout{** the default language instead.}%
\else
\language=\csname l@#1\endcsname
\fi
#2}}
\providecommand{\BIBdecl}{\relax}
\BIBdecl

\bibitem[Jones and Lederman(2006)]{jones2006human}
L.~A. Jones and S.~J. Lederman, \emph{Human hand function}.\hskip 1em plus
  0.5em minus 0.4em\relax Oxford University Press, 2006.

\bibitem[Norrsell et~al.(1999)Norrsell, Finger, and
  Lajonchere]{norrsell1999cutaneous}
U.~Norrsell, S.~Finger, and C.~Lajonchere, ``Cutaneous sensory spots and the
  “law of specific nerve energies”: history and development of ideas,''
  \emph{Brain research bulletin}, vol.~48, no.~5, pp. 457--465, 1999.

\bibitem[Donlon et~al.(2018)Donlon, Dong, Liu, Li, Adelson, and
  Rodriguez]{donlon2018gelslim}
E.~Donlon, S.~Dong, M.~Liu, J.~Li, E.~Adelson, and A.~Rodriguez, ``Gelslim: A
  high-resolution, compact, robust, and calibrated tactile-sensing finger,''
  \emph{IEEE/RSJ International Conference on Intelligent Robots and Systems
  (IROS)}, 2018.

\bibitem[Yuan et~al.(2017{\natexlab{a}})Yuan, Dong, and
  Adelson]{GelSight_review}
W.~Yuan, S.~Dong, and E.~H. Adelson, ``Gelsight: High-resolution robot tactile
  sensors for estimating geometry and force,'' \emph{Sensors}, vol.~17, no.~12,
  p. 2762, 2017.

\bibitem[Dong et~al.(2018)Dong, Ma, Donlon, and Rodriguez]{dong2019}
\BIBentryALTinterwordspacing
S.~Dong, D.~Ma, E.~Donlon, and A.~Rodriguez, ``Maintaining grasps within
  slipping bound by monitoring incipient slip,'' \emph{CoRR}, vol.
  abs/1810.13381, 2018. [Online]. Available:
  \url{http://arxiv.org/abs/1810.13381}
\BIBentrySTDinterwordspacing

\bibitem[{Ma} et~al.(2018){Ma}, {Donlon}, {Dong}, and {Rodriguez}]{ma2019}
D.~{Ma}, E.~{Donlon}, S.~{Dong}, and A.~{Rodriguez}, ``{Dense Tactile Force
  Distribution Estimation using GelSlim and inverse FEM},'' \emph{arXiv
  e-prints}, p. arXiv:1810.04621, Oct 2018.

\bibitem[Newcombe et~al.(2011)Newcombe, Izadi, Hilliges, Molyneaux, Kim,
  Davison, Kohi, Shotton, Hodges, and Fitzgibbon]{newcombe2011}
R.~A. Newcombe, S.~Izadi, O.~Hilliges, D.~Molyneaux, D.~Kim, A.~J. Davison,
  P.~Kohi, J.~Shotton, S.~Hodges, and A.~Fitzgibbon, ``Kinectfusion: Real-time
  dense surface mapping and tracking,'' in \emph{2011 10th IEEE International
  Symposium on Mixed and Augmented Reality}, Oct 2011, pp. 127--136.

\bibitem[Salas-Moreno et~al.(2013)Salas-Moreno, Newcombe, Strasdat, Kelly, and
  Davison]{Salas-Moreno2013}
R.~F. Salas-Moreno, R.~A. Newcombe, H.~Strasdat, P.~H. Kelly, and A.~J.
  Davison, ``Slam++: Simultaneous localisation and mapping at the level of
  objects,'' in \emph{The IEEE Conference on Computer Vision and Pattern
  Recognition (CVPR)}, June 2013.

\bibitem[OpenAI et~al.(2018)OpenAI, Andrychowicz, Baker, Chociej, Jozefowicz,
  McGrew, Pachocki, Petron, Plappert, Powell, Ray, Schneider, Sidor, Tobin,
  Welinder, Weng, and Zaremba]{OpenAI2018}
OpenAI, M.~Andrychowicz, B.~Baker, M.~Chociej, R.~Jozefowicz, B.~McGrew,
  J.~Pachocki, A.~Petron, M.~Plappert, G.~Powell, A.~Ray, J.~Schneider,
  S.~Sidor, J.~Tobin, P.~Welinder, L.~Weng, and W.~Zaremba, ``Learning
  dexterous in-hand manipulation,'' 2018.

\bibitem[Björkman et~al.(2013)Björkman, Bekiroglu, Högman, and
  Kragic]{Bjorkman2013}
M.~Björkman, Y.~Bekiroglu, V.~Högman, and D.~Kragic, ``Enhancing visual
  perception of shape through tactile glances,'' in \emph{2013 IEEE/RSJ
  International Conference on Intelligent Robots and Systems}, Nov 2013, pp.
  3180--3186.

\bibitem[Allen et~al.(1999)Allen, Miller, Oh, and Leibowitz]{Allen1999}
P.~K. Allen, A.~T. Miller, P.~Y. Oh, and B.~S. Leibowitz, ``Integration of
  vision, force and tactile sensing for grasping,'' \emph{Int. J. Intelligent
  Machines}, vol.~4, pp. 129--149, 1999.

\bibitem[Ilonen et~al.(2014)Ilonen, Bohg, and Kyrki]{Ilonen2014}
\BIBentryALTinterwordspacing
J.~Ilonen, J.~Bohg, and V.~Kyrki, ``Three-dimensional object reconstruction of
  symmetric objects by fusing visual and tactile sensing,'' \emph{The
  International Journal of Robotics Research}, vol.~33, no.~2, pp. 321--341,
  2014. [Online]. Available: \url{https://doi.org/10.1177/0278364913497816}
\BIBentrySTDinterwordspacing

\bibitem[Luo et~al.(2017{\natexlab{a}})Luo, Mou, Althoefer, and Liu]{Luo2017}
\BIBentryALTinterwordspacing
S.~Luo, W.~Mou, K.~Althoefer, and H.~Liu, ``Localizing the object contact
  through matching tactile features with visual map,'' \emph{CoRR}, vol.
  abs/1708.04441, 2017. [Online]. Available:
  \url{http://arxiv.org/abs/1708.04441}
\BIBentrySTDinterwordspacing

\bibitem[Falco et~al.(2017)Falco, Lu, Cirillo, Natale, Pirozzi, and
  Lee]{Falco2017}
P.~Falco, S.~Lu, A.~Cirillo, C.~Natale, S.~Pirozzi, and D.~Lee, ``Cross-modal
  visuo-tactile object recognition using robotic active exploration,'' in
  \emph{Robotics and Automation (ICRA), 2017 IEEE International Conference
  on}.\hskip 1em plus 0.5em minus 0.4em\relax IEEE, 2017, pp. 5273--5280.

\bibitem[Strub et~al.(2014)Strub, Wörgötter, Ritter, and
  Sandamirskaya]{Strub2014}
C.~Strub, F.~Wörgötter, H.~Ritter, and Y.~Sandamirskaya, ``Using haptics to
  extract object shape from rotational manipulations,'' in \emph{2014 IEEE/RSJ
  International Conference on Intelligent Robots and Systems}, Sept 2014, pp.
  2179--2186.

\bibitem[Luo et~al.(2018)Luo, Mou, Althoefer, and Liu]{Luo2018}
S.~Luo, W.~Mou, K.~Althoefer, and H.~Liu, ``iclap: shape recognition by
  combining proprioception and touch sensing,'' \emph{Autonomous Robots}, pp.
  1--12, 2018.

\bibitem[Martinez-Hernandez et~al.(2013)Martinez-Hernandez, Metta, Dodd,
  Prescott, Natale, and Lepora]{Martinez-Hernandez2013}
U.~Martinez-Hernandez, G.~Metta, T.~J. Dodd, T.~J. Prescott, L.~Natale, and
  N.~F. Lepora, ``Active contour following to explore object shape with robot
  touch,'' in \emph{2013 World Haptics Conference (WHC)}, April 2013, pp.
  341--346.

\bibitem[Jamali et~al.(2016)Jamali, Ciliberto, Rosasco, and Natale]{Jamali2016}
N.~Jamali, C.~Ciliberto, L.~Rosasco, and L.~Natale, ``Active perception:
  Building objects' models using tactile exploration,'' in \emph{2016 IEEE-RAS
  16th International Conference on Humanoid Robots (Humanoids)}, Nov 2016, pp.
  179--185.

\bibitem[Yi et~al.(2016)Yi, Calandra, Veiga, van Hoof, Hermans, Zhang, and
  Peters]{Yi2016}
Z.~Yi, R.~Calandra, F.~Veiga, H.~van Hoof, T.~Hermans, Y.~Zhang, and J.~Peters,
  ``Active tactile object exploration with gaussian processes,'' in
  \emph{Intelligent Robots and Systems (IROS), 2016 IEEE/RSJ International
  Conference on}.\hskip 1em plus 0.5em minus 0.4em\relax IEEE, 2016, pp.
  4925--4930.

\bibitem[Driess et~al.(2017)Driess, Englert, and Toussaint]{Driess2017}
D.~Driess, P.~Englert, and M.~Toussaint, ``Active learning with query paths for
  tactile object shape exploration,'' in \emph{Intelligent Robots and Systems
  (IROS), 2017 IEEE/RSJ International Conference on}.\hskip 1em plus 0.5em
  minus 0.4em\relax IEEE, 2017, pp. 65--72.

\bibitem[Sommer et~al.(2014)Sommer, Li, and Billard]{Sommer2014}
N.~Sommer, M.~Li, and A.~Billard, ``Bimanual compliant tactile exploration for
  grasping unknown objects,'' in \emph{2014 IEEE International Conference on
  Robotics and Automation (ICRA)}, May 2014, pp. 6400--6407.

\bibitem[Zhang et~al.(2017)Zhang, Atanasov, and Daniilidis]{zhang2017active}
M.~M. Zhang, N.~Atanasov, and K.~Daniilidis, ``Active end-effector pose
  selection for tactile object recognition through monte carlo tree search,''
  in \emph{2017 IEEE/RSJ International Conference on Intelligent Robots and
  Systems (IROS)}.\hskip 1em plus 0.5em minus 0.4em\relax IEEE, 2017, pp.
  3258--3265.

\bibitem[Mao et~al.(2017)Mao, Xiao, Zhang, and Daniilidis]{mao2017shape}
H.~Mao, J.~Xiao, M.~M. Zhang, and K.~Daniilidis, ``Shape-based object
  classification and recognition through continuum manipulation,'' in
  \emph{2017 IEEE/RSJ International Conference on Intelligent Robots and
  Systems (IROS)}.\hskip 1em plus 0.5em minus 0.4em\relax IEEE, 2017, pp.
  456--463.

\bibitem[Luo et~al.(2017{\natexlab{b}})Luo, Bimbo, Dahiya, and
  Liu]{Luo2017review}
S.~Luo, J.~Bimbo, R.~Dahiya, and H.~Liu, ``Robotic tactile perception of object
  properties: A review,'' \emph{Mechatronics}, vol.~48, pp. 54--67, 2017.

\bibitem[Petrovskaya and Khatib(2011)]{petrovskaya2011}
A.~Petrovskaya and O.~Khatib, ``Global localization of objects via touch,''
  \emph{IEEE Transactions on Robotics}, vol.~27, no.~3, pp. 569--585, 2011.

\bibitem[Koval et~al.(2015)Koval, Pollard, and Srinivasa]{koval2015}
M.~C. Koval, N.~S. Pollard, and S.~S. Srinivasa, ``Pose estimation for planar
  contact manipulation with manifold particle filters,'' \emph{The
  International Journal of Robotics Research}, vol.~34, no.~7, pp. 922--945,
  2015.

\bibitem[Moll and Erdmann(2004)]{moll2004}
M.~Moll and M.~A. Erdmann, ``Reconstructing the shape and motion of unknown
  objects with active tactile sensors,'' in \emph{Algorithmic Foundations of
  Robotics V}.\hskip 1em plus 0.5em minus 0.4em\relax Springer, 2004, pp.
  293--309.

\bibitem[Pezzementi et~al.(2011)Pezzementi, Reyda, and Hager]{pezzementi2011}
Z.~Pezzementi, C.~Reyda, and G.~D. Hager, ``Object mapping, recognition, and
  localization from tactile geometry,'' in \emph{Robotics and Automation
  (ICRA), 2011 IEEE International Conference on}.\hskip 1em plus 0.5em minus
  0.4em\relax IEEE, 2011, pp. 5942--5948.

\bibitem[Ottenhaus et~al.(2016)Ottenhaus, Miller, Schiebener, Vahrenkamp, and
  Asfour]{ottenhaus2016}
S.~Ottenhaus, M.~Miller, D.~Schiebener, N.~Vahrenkamp, and T.~Asfour, ``Local
  implicit surface estimation for haptic exploration,'' in \emph{Humanoid
  Robots (Humanoids), 2016 IEEE-RAS 16th International Conference on}.\hskip
  1em plus 0.5em minus 0.4em\relax IEEE, 2016, pp. 850--856.

\bibitem[Aggarwal and Kirchner(2014)]{aggarwal2014}
A.~Aggarwal and F.~Kirchner, ``Object recognition and localization: The role of
  tactile sensors,'' \emph{Sensors}, vol.~14, no.~2, pp. 3227--3266, 2014.

\bibitem[Besl and McKay(1992)]{besl1992}
P.~J. Besl and N.~D. McKay, ``Method for registration of 3-d shapes,'' in
  \emph{Sensor Fusion IV: Control Paradigms and Data Structures}, vol.
  1611.\hskip 1em plus 0.5em minus 0.4em\relax International Society for Optics
  and Photonics, 1992, pp. 586--607.

\bibitem[Dellaert and Kaess(2006)]{dellaert2006}
F.~Dellaert and M.~Kaess, ``Square root sam: Simultaneous localization and
  mapping via square root information smoothing,'' \emph{The International
  Journal of Robotics Research}, vol.~25, no.~12, pp. 1181--1203, 2006.

\bibitem[Yu et~al.(2015)Yu, Leonard, and Rodriguez]{Yu2015}
K.-T. Yu, J.~Leonard, and A.~Rodriguez, ``{Shape and Pose Recovery from Planar
  Pushing},'' in \emph{IROS}, 2015.

\bibitem[Yuan et~al.(2017{\natexlab{b}})Yuan, Zhu, Owens, Srinivasan, , and
  Adelson]{yuan_2017}
W.~Yuan, C.~Zhu, A.~Owens, M.~A. Srinivasan, , and E.~H. Adelson,
  ``Shape-independent hardness estimation using deep learning and a gelsight
  tactile sensor,'' in \emph{IEEE International Conference on Robotics and
  Automation (ICRA)}, 2017.

\bibitem[Li et~al.(2014)Li, Platt, Yuan, ten Pas, Roscup, Srinivasan, and
  Adelson]{li_2014}
R.~Li, R.~Platt, W.~Yuan, A.~ten Pas, N.~Roscup, M.~A. Srinivasan, and
  E.~Adelson, ``Localization and manipulation of small parts using gelsight
  tactile sensing,'' in \emph{Intelligent Robots and Systems (IROS), 2014
  IEEE/RSJ International Conference on.}, 2014.

\bibitem[Izatt et~al.(2017)Izatt, Mirano, Adelson, and Tedrake]{izatt_2017}
G.~Izatt, G.~Mirano, E.~Adelson, and R.~Tedrake, ``Tracking objects with point
  clouds from vision and touch,'' in \emph{IEEE International Conference on
  Robotics and Automation (ICRA)}, 2017, pp. 4000--4007.

\bibitem[{Wang} et~al.(2018){Wang}, {Wu}, {Sun}, {Yuan}, {Freeman},
  {Tenenbaum}, and {Adelson}]{Wang2018}
S.~{Wang}, J.~{Wu}, X.~{Sun}, W.~{Yuan}, W.~T. {Freeman}, J.~B. {Tenenbaum},
  and E.~H. {Adelson}, ``{3D Shape Perception from Monocular Vision, Touch, and
  Shape Priors},'' \emph{ArXiv e-prints}, Aug. 2018.

\bibitem[Zeng et~al.(2018)Zeng, Song, Yu, et~al.]{zeng_2017}
A.~Zeng, S.~Song, K.~Yu \emph{et~al.}, ``Robotic pick-and-place of novel
  objects in clutter with multi-affordance grasping and cross-domain image
  matching,'' in \emph{IEEE International Conference on Robotics and Automation
  (ICRA)}, 2018.

\bibitem[Hogan et~al.(2018)Hogan, Bauz{\'a}, Canal, Donlon, and
  Rodriguez]{Hogan2018}
F.~R. Hogan, M.~Bauz{\'a}, O.~Canal, E.~Donlon, and A.~Rodriguez, ``Tactile
  regrasp: Grasp adjustments via simulated tactile transformations,''
  \emph{CoRR}, vol. abs/1803.01940, 2018.

\bibitem[pro()]{projectwebsite}
\BIBentryALTinterwordspacing
Website for push dataset. [Online]. Available:
  \url{web.mit.edu/mcube/research/tactile_localization}
\BIBentrySTDinterwordspacing

\bibitem[He et~al.(2016)He, Zhang, Ren, and Sun]{he2016}
K.~He, X.~Zhang, S.~Ren, and J.~Sun, ``Deep residual learning for image
  recognition,'' in \emph{IEEE Conference on Computer Vision and Pattern
  Recognition (CVPR)}, 2016.

\end{thebibliography}

\end{document}